\documentclass[lettersize,journal]{IEEEtran}
\usepackage{amsmath,amsfonts}
\usepackage{array}
\usepackage[caption=false,font=normalsize,labelfont=sf,textfont=sf]{subfig}
\usepackage{textcomp}
\usepackage{stfloats}
\usepackage{url}
\usepackage{verbatim}
\usepackage{graphicx}
\usepackage{cite}
\usepackage{multirow}
\usepackage{booktabs}
\usepackage[table]{xcolor}
\usepackage[normalem]{ulem}
\useunder{\uline}{\ul}{}
\usepackage[linesnumbered,ruled,vlined]{algorithm2e}
\usepackage[colorlinks, linkcolor=blue, anchorcolor=blue, citecolor=blue]{hyperref}

\begin{document}

\title{Zero-Query Transferable Pretend Benign Attacks with Amortized Online Generation \\ for Cooperative Perception}

\author{Hongwei~Lin, Xun~Huang, 
Chenglu~Wen$^\dag$,~\IEEEmembership{Senior~Member,~IEEE},
and~Cheng~Wang,~\IEEEmembership{Senior~Member,~IEEE}
        
\thanks{Hongwei~Lin, Xun~Huang, Chenglu~Wen, and Cheng~Wang are with the Fujian Key Laboratory of Urban Intelligent Sensing and Computing and the School of Informatics, Xiamen University, Xiamen, FJ 361005, China. (E-mail: \{greatlin, huangxun\}@stu.xmu.edu.cn; \{clwen, cwang\}@xmu.edu.cn). Hongwei~Lin and Xun~Huang are also with Zhongguancun Academy, Beijing, China.}
\thanks{$^\dag$ Corresponding author.}}



\maketitle

\begin{abstract}
Cooperative perception enables an ego agent to acquire information from collaborators beyond the reach of its own sensors. Yet, the messages supplying the greatest benefit are precisely those the ego agent is least able to verify from its own observations. Pretend Benign (PB) exploits this asymmetry by selecting regions of weak local verifiability and imposing multi-objective perturbation, thereby making harmful feature changes appear as plausible benign assistance. However, this mechanism is victim-dependent: PB requires the victim model structure, parameters, and gradients for region selection and perturbation optimization, limiting direct transfer to unseen victims. To eliminate this dependence, the attack must generalize across victims, so that the perturbations it produces reflect properties shared by cooperative detection in general instead of those specific to any single network. Building on this principle, this work constructs a victim-agnostic attack capability from a heterogeneous ensemble of surrogate models, and consolidates this capability into a reusable dictionary that amortizes the per-message cost of attack generation. Specifically, Transferable Pretend Benign (TrPB) transfers PB's region selection criterion and multi-objective perturbation strategy across heterogeneous surrogate models to construct a victim-agnostic attack offline. We further design RT-TrPB to amortize the per-message cost of TrPB by encoding its transferable regions and perturbation patterns into a region predictor and reusable perturbation dictionary. On widely used cooperative perception benchmarks, RT-TrPB bypasses the consensus-based defenses ROBOSAC and FLD at rates approaching 100\% and generates each perturbation in under 56 ms. To counter PB-style attacks that masquerade as benign collaboration, we further design STTrust to verify each received message against spatial and temporal consistency and to weight the verification outcome by identity-level trust. STTrust detects over 90\% of PB-style attacks and restores perception performance close to the attack-free level. The source code will be publicly available at
\url{https://github.com/windlinsherlock/Pretend-Benign}.
\end{abstract}


\section{Introduction}
\label{sec:intro}

\begin{figure}[t]
  \centering
   \includegraphics[width=\linewidth]{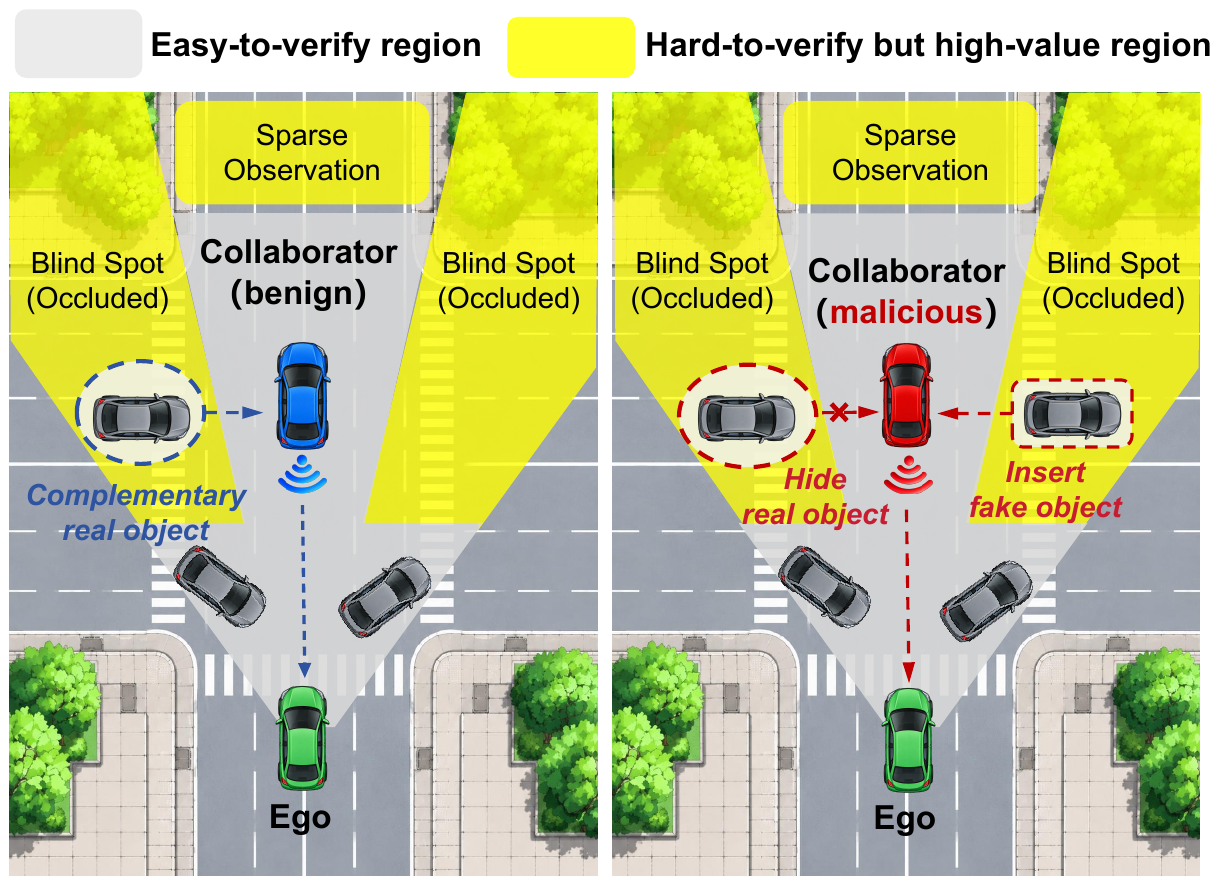}
   \caption{Verification asymmetry in cooperative perception. Collaborator messages provide high-value evidence in regions outside the ego vehicle's reliable view, but those regions are difficult to verify independently. Pretend Benign (PB)~\cite{pb} exploits this asymmetry by preserving agreement in verifiable regions while manipulating complementary evidence.}
   \label{fig:verification_asymmetry}
\end{figure}

\IEEEPARstart{C}{ooperative} perception extends an ego vehicle's sensing coverage by integrating complementary observations from collaborating agents, enabling the perception of objects that are occluded from or beyond the sensing range of the ego vehicle~\cite{F-cooper, v2vnet, when2com, opv2v, v2xset}. The fundamental benefit of collaboration lies in accessing scene evidence unavailable from the ego vehicle's local observations alone.
However, information received from collaborators may be falsified during transmission, making consistency verification against the ego vehicle's own observations a critical safeguard prior to fusion. The effectiveness of such verification, however, is inherently limited in complementary regions, where the ego vehicle has sparse observations and thus lacks sufficient evidence for reliable cross-checking. This creates a verification asymmetry: the information that offers the greatest collaborative value is also the information that the ego vehicle is least capable of independently verifying, as illustrated in Fig.~\ref{fig:verification_asymmetry}.

However, existing attacks have largely overlooked the verification asymmetry inherent in cooperative perception. They typically perturb transmitted features and poses or inject fabricated perception messages~\cite{James_attack, advgps, data-attacks}, without considering whether the manipulated regions are supported by the ego vehicle's own observations. Although such attacks can substantially degrade the victim's perception, discrepancies with ego-side evidence may be exposed by consistency checks~\cite{robosac, made}, limiting their stealth. To overcome this limitation, our prior work introduced Pretend Benign (PB)~\cite{pb}, an attack strategy that explicitly exploits the verification asymmetry. PB preserves consistency with ego-side observations in regions where reliable local evidence is available, while concentrating malicious feature perturbations in complementary regions that the ego vehicle cannot independently verify. Consequently, the manipulated message can maintain apparent consistency with the ego vehicle's observations while introducing errors into downstream perception.

\begin{figure}[t]
  \centering
   \includegraphics[width=\linewidth]{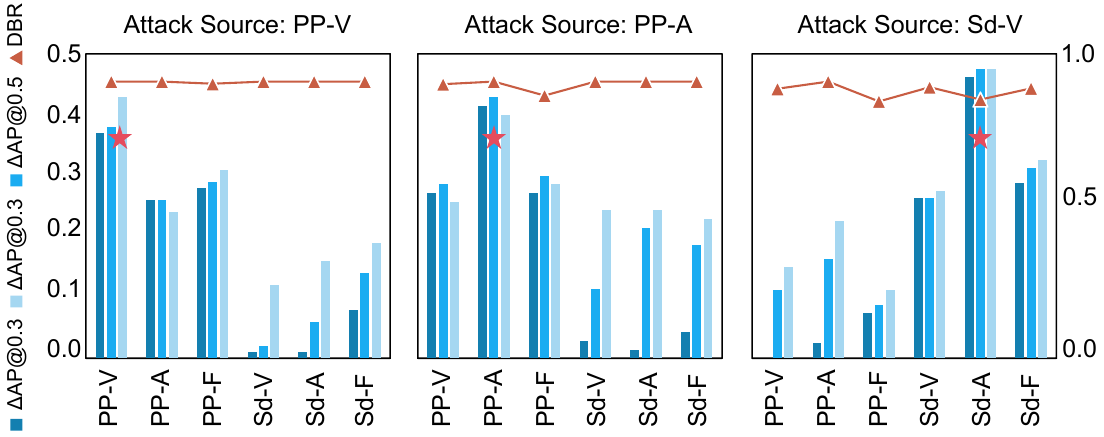}
   \caption{Transferability of prior PB across heterogeneous cooperative perception models. Bars report AP degradation at three IoU thresholds, and lines report the defense-bypass rate (DBR) for perturbations generated from each source model. The variation across source--victim model pairs shows that victim-dependent PB construction does not reliably transfer across heterogeneous backbones.}
   \label{fig:PB_transfer}
\end{figure}

The original formulation of PB~\cite{pb} is a white-box attack that selects attack regions and optimizes perturbations using victim-specific model structure, parameters, and gradients. As further demonstrated in Fig.~\ref{fig:PB_transfer}, PB perturbations optimized for one victim transfer poorly to unseen heterogeneous victims, particularly when their backbones differ. This reveals a key limitation of the prior attack: its region-specific stealth behavior cannot be directly reused across heterogeneous victims and must be reconstructed for each target. Although existing work has begun to investigate cross-model transfer in cooperative perception~\cite{James_attack,SOMBRA}, it has not yet established a transferable black-box attack that addresses this limitation. The key unresolved question is whether PB's region-specific stealth objective and benign-consistent perturbation behavior can be transferred to an unseen heterogeneous victim under a zero-query setting. Here, zero-query refers to attack construction without access to the victim's model structure, parameters, gradients, or prediction feedback.

\begin{figure*}[t]
  \centering
   \includegraphics[width=0.85\linewidth]{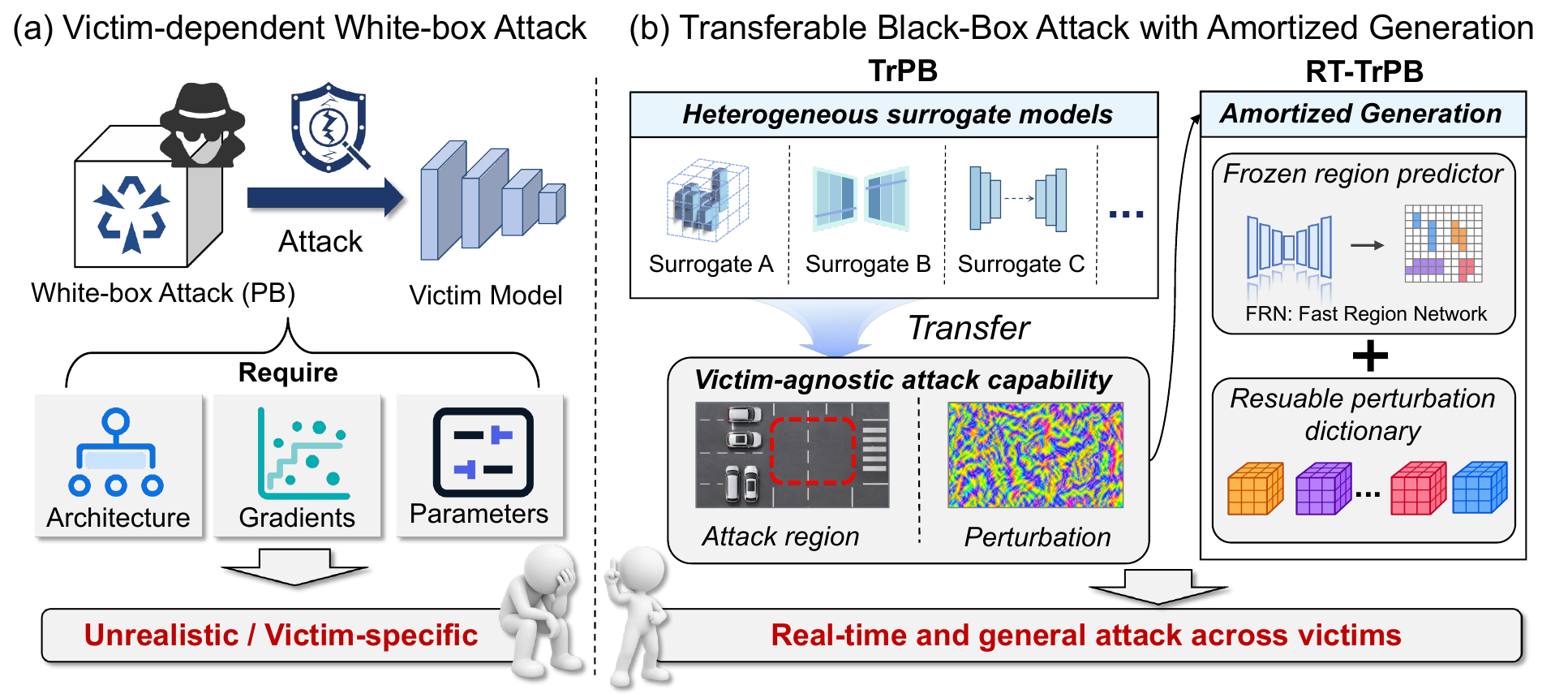}
   \caption{From victim-dependent PB construction to transferable and amortized online generation. (a) Conventional PB \cite{pb} relies on victim-specific access to the model architecture, gradients, and parameters, making it impractical for black-box attacks; (b) TrPB aggregates knowledge from heterogeneous surrogate models to construct transferable, victim-agnostic attack regions and perturbation patterns under the zero-query setting, while RT-TrPB amortizes their online generation through a frozen region predictor and a reusable perturbation dictionary, enabling efficient, real-time black-box attacks across different victims without per-frame region aggregation or iterative perturbation optimization.}
   \label{fig:TrPB_motivation}
\end{figure*}

Within this zero-query setting, we introduce Transferable Pretend Benign (TrPB) and RT-TrPB to address two successive bottlenecks: TrPB makes PB's region-specific stealth behavior transferable across heterogeneous victims, while RT-TrPB amortizes the resulting knowledge for efficient online attack generation. The offline-to-online pipeline is illustrated in Fig.~\ref{fig:TrPB_motivation}. TrPB constructs reusable knowledge across heterogeneous surrogate models through two steps: first, aggregating surrogate-derived PB regions to identify complementary areas with limited ego-side evidence; and then jointly optimizing shared, region-constrained perturbations that preserve PB's benign-consistency constraints while altering downstream perception. This ordering reduces dependence on model-specific region selections and favors perturbation patterns that generalize across surrogate objectives. RT-TrPB further amortizes this offline knowledge into a region predictor and a perturbation dictionary. The predictor selects PB regions based on the current cooperative context, while the dictionary retrieves compatible perturbation patterns for single-pass composition, avoiding per-instance iterative optimization. Together, TrPB and RT-TrPB establish an offline-to-online progression: TrPB constructs transferable attack knowledge across surrogate models, while RT-TrPB enables its efficient online generation.

\begin{figure}[t]
  \centering
   \includegraphics[width=\linewidth]{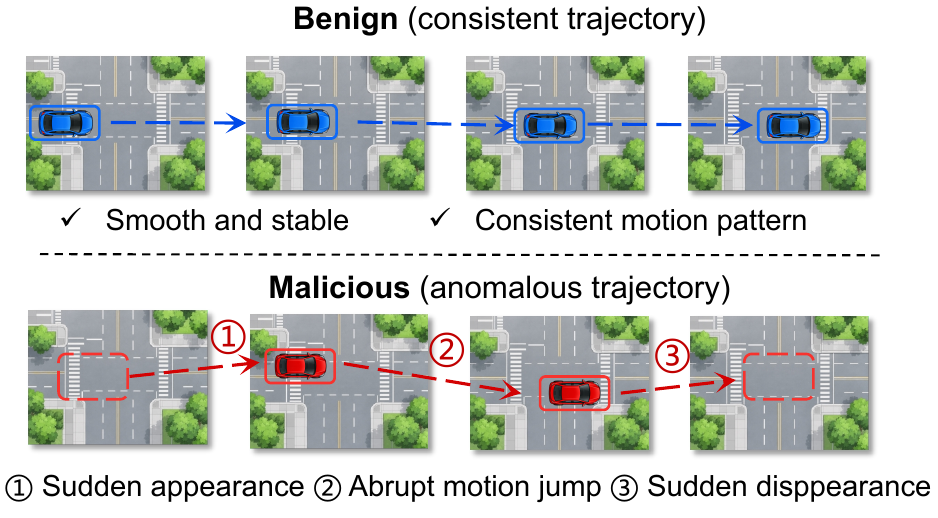}
   \caption{Motivation for spatio-temporal trust verification. A single frame may preserve local spatial agreement, whereas temporal inconsistencies can expose anomalous appearance, disappearance, or motion jumps, motivating spatio-temporal consistency checks.}
   \label{fig:STTrust_motivation}
\end{figure}

The preceding attack analysis highlights a broader defense requirement: a verifier should detect both attacks that introduce observable inconsistencies and those that preserve local agreement. Existing defenses, including ROBOSAC~\cite{robosac}, FLD~\cite{pb}, and MADE~\cite{made}, rely on single-frame spatial consistency checks, while GCP~\cite{GCP}, PRBI~\cite{PRBI}, and MATE~\cite{Hallyburton2025MATE} further incorporate spatio-temporal consistency or historical trust cues to detect conventional feature- and pose-level attacks that introduce discrepancies with ego-side evidence~\cite{James_attack, advgps}. However, these defenses remain insufficient against attacks that preserve local consistency while manipulating complementary regions beyond the ego vehicle's independent verification capability, such as PB~\cite{pb}. To address this challenge, we develop the Spatio-Temporal Trust framework (STTrust), a strengthened spatio-temporal consistency-based verifier, as illustrated in Fig.~\ref{fig:STTrust_motivation}. STTrust retains spatial consistency checks for conventional attacks and augments them with motion-compensated temporal evidence and identity-level trust, evaluating single-frame agreement against evidence accumulated across frames and collaborators.

Building on our prior work on PB~\cite{pb}, this paper extends the study in three directions: TrPB enables zero-query cross-victim transfer, RT-TrPB supports amortized online attack generation, and STTrust provides spatio-temporal verification. The proposed methods are evaluated on simulated and real-world cooperative perception datasets, including OPV2V~\cite{opv2v}, V2XSet~\cite{v2xset}, and V2V4Real~\cite{V2v4real}. The main contributions are summarized as follows:

\begin{itemize}
\item We extend PB beyond victim-dependent white-box construction by formulating and validating zero-query transfer of its region-specific stealth behavior to unseen victims under compatible communicated feature interfaces.
\item We unify transfer and deployment through a shared attack-knowledge path: TrPB constructs cross-model region semantics and constrained perturbations offline, while RT-TrPB amortizes them into single-pass online generation.
\item We develop STTrust to extend single-frame spatial consistency checking with motion-compensated temporal evidence and identity-level trust, enabling the detection of both conventional and PB-style attacks.
\item Across the evaluated benchmarks, RT-TrPB reaches up to 100\% defense-bypass rates against the evaluated defenses with 40.18--55.54~ms perturbation-generation latency, while STTrust rejects more than 90\% of PB-style attacks and substantially restores degraded perception.
\end{itemize}
\section{Related Work}
\label{sec:related}

\subsection{Multi-Agent Cooperative Perception}
Multi-agent cooperative perception (CP) extends an agent's sensing capability by exchanging and fusing complementary observations. By combining feature- or message-level evidence from different viewpoints, CP can cover occluded objects and regions outside the ego sensor range. Representative systems include F-Cooper~\cite{F-cooper}, V2VNet~\cite{v2vnet}, When2com~\cite{when2com}, and V2X-ViT~\cite{v2xset}, which use intermediate features, learned communication, adaptive grouping, or transformer fusion. OPV2V~\cite{opv2v} and V2XSet~\cite{v2xset} support simulated evaluation, while V2V4Real~\cite{V2v4real} provides a real-world benchmark. Subsequent work addresses communication efficiency, temporal awareness, heterogeneous modalities, compact messages, sequential perception, and robustness to natural corruptions~\cite{how2comm, scope, linking, pragmatic, CoLC, discoNet, core, hm-vit, di-v2x, MRCnet, codefilling, Point-cluster, CoopTrack, Cosdh, V2x-vitv2, pragmatic2, CoopDiff}. These efforts improve what is communicated and how it is fused, but leave a security dependency: when a collaborator is useful precisely because it supplies evidence absent from the ego view, the receiver cannot reconstruct that evidence locally. This verification asymmetry makes the communication layer a distinct security surface for the attacks and defenses reviewed next.

\subsection{Adversarial Attacks and Defenses in Cooperative Perception}
Because CP's benefit depends on accepting external evidence, a compromised collaborator can alter the receiver's perception before the ego vehicle can fully verify what that collaborator contributes. Security research asks how attacks alter perception and how verification can distinguish them from legitimate complementary observations. James' Attack~\cite{James_attack} perturbs intermediate representations, AdvGPS~\cite{advgps} manipulates cooperative pose information, and data-fabrication work~\cite{data-attacks} studies forged messages. PosePert~\cite{PosePert} and TrustFlip~\cite{TrustFlip} extend the surface to realistic trajectory and physical-scene effects. These attacks often create discrepancies that consistency checks can expose. PB~\cite{pb} changes the objective by exploiting verification asymmetry, modifying weakly verifiable regions while preserving benign-like collaboration. SOMBRA~\cite{SOMBRA} evaluates white-box source-model perturbations on other destinations, while MVIG~\cite{MVIG} studies model-aware adaptive attacks using mutual-view and temporal information. The unresolved problem is to transfer PB-specific region semantics and benign-consistency constraints to an unseen heterogeneous victim without victim-model information or feedback. 

Defenses can be organized by the evidence they trust. ROBOSAC~\cite{robosac}, MADE~\cite{made}, and PB's FLD~\cite{pb} compare collaborators through consensus or feature-level consistency. Latent Guardian/Cerberus~\cite{LatentGuardian}, GLST~\cite{GLST}, and FNO-Guard~\cite{FNO-Guard} further inspect learned feature structure. GCP~\cite{GCP}, PRBI~\cite{PRBI}, CP-UniGuard~\cite{CP-uniGuard}, and MATE~\cite{Hallyburton2025MATE} add spatio-temporal, historical, or identity-level evidence, while ATConsensus~\cite{ATConsensus} and LUCIA~\cite{SOMBRA} combine robustness training or trust-aware fusion with detection. This progression motivates STTrust as a spatio-temporal consistency verifier that must handle both discrepancy-producing attacks and PB-style attacks while preserving benign collaboration.

\subsection{Black-box Adversarial Attacks}
The central distinction in black-box research is what the attacker can observe from the victim. Query-based methods such as NES~\cite{NES}, Square Attack~\cite{Square_Attack}, HSJA~\cite{HSJA}, and later low-query variants~\cite{black-query-1, black-query-2} use score, decision, or boundary feedback. Transfer-based methods remove that interaction by constructing perturbations on surrogate models~\cite{Explaining, transferability}. Momentum, input diversity, translation invariance, ensemble diversity, and model/checkpoint diversification improve transfer, but recent 3D results also show that success depends on the representation and model family~\cite{transfer-1, transfer-2, black-transfer-1, black-transfer-2, black-transfer-3}. Cooperative perception sharpens this problem because feature messages are model-dependent and a useful attack must preserve the appearance of legitimate collaboration, rather than merely optimize task loss. James' Attack~\cite{James_attack} still adapts to victim-side feature observations, while SOMBRA~\cite{SOMBRA} transfers perturbations constructed with a white-box source model. Unlike these generic or victim-adapted transfer mechanisms, our proposed TrPB transfers PB-specific region semantics and benign-consistency constraints across heterogeneous surrogates. Our proposed RT-TrPB then amortizes this shared attack knowledge into single-stage online generation under a zero-query message interface.

\section{Problem Formulation}
\label{sec:problem_pb}

This section defines the intermediate-feature cooperative perception interface, the information boundary for white-box and zero-query attacks, and the attacker and defender objectives used throughout the paper. We first specify what each agent computes and transmits, then describe how an injected feature changes the victim prediction and how the defense forms its accepted set. The resulting notation is shared by the PB formulation and the later transferable construction.

\subsection{Cooperative Perception and Observable Information}
\label{sec:pre}

We consider a cooperative perception system with agents $\mathcal{V}=\{1,\ldots,N\}$. Agent $i$ observes its local input $x_i$ and extracts an intermediate bird's-eye-view (BEV) feature $f_i^{\mathrm{raw}}=\mathcal{E}_i(x_i)$, where $\mathcal{E}_i$ is the local feature encoder. Let $e$ denote the ego agent and let $\mathcal{N}_e=\mathcal{V}\setminus\{e\}$ denote its collaborators. The ego agent aligns the feature of collaborator $j\in\mathcal{N}_e$ to the ego coordinate system before fusion as:
\begin{equation}
f_{j\rightarrow e}=\mathrm{Align}_{j\rightarrow e}(f_j^{\mathrm{raw}}).
\label{eq:aligned_feature}
\end{equation}
The communicated message contains this aligned intermediate feature, the agent pose, and the capture timestamp. We write it as $m_{j\rightarrow e}=\left(f_{j\rightarrow e},p_j,t_j\right)$, where $p_j$ is the pose of agent $j$ and $t_j$ is the timestamp associated with the feature. No raw point cloud or detector output is required for this message interface. This intermediate-feature interface is representative of feature-level cooperative perception systems~\cite{F-cooper,v2vnet,V2x-vitv2}. For notational simplicity, after alignment we write $f_e=f_e^{\mathrm{raw}}$ and $f_j=f_{j\rightarrow e}$ whenever the coordinate system is clear.

The ego agent fuses its local feature and the received collaborator features into a shared BEV representation. The detection head then utilizes this representation to generate scene perception predictions:
\begin{equation}
\begin{aligned}
z_e&=\mathcal{F}_e\!\left(f_e,\{f_j\}_{j\in\mathcal{N}_e}\right),\\
Y_e&=\mathcal{P}_e(z_e)=\left\{y_e^{(k)}\right\}_{k=1}^{K_e}.
\end{aligned}
\label{eq:cp_interface}
\end{equation}
where $\mathcal{F}_e$ denotes the fusion module. Here, $Y_e$ is the set of predicted objects, and each $y_e^{(k)}$ contains a 3D bounding box $b_e^{(k)}$, an object category $c_e^{(k)}$, and a confidence score $s_e^{(k)}$. The classification branch is denoted by $\mathcal{P}_{\mathrm{cls}}$. Its response maps are
\begin{equation}
\begin{aligned}
h_e&=\mathcal{P}_{\mathrm{cls}}(f_e),\qquad
h_a=\mathcal{P}_{\mathrm{cls}}(f_a),\qquad
h_z=\mathcal{P}_{\mathrm{cls}}(z_e).
\end{aligned}
\label{eq:pb_heatmaps}
\end{equation}
Based on these response maps, the corresponding foreground supports are obtained as $\Omega_i^{+}=\Phi(h_i)$ for $i\in\{e,a,z\}$, where $\Phi(\cdot)$ decodes foreground support on the complete BEV lattice $\Omega$. These response maps and their associated supports are subsequently used to identify the PB regions.

The attacker observes only the information exposed by the message protocol. We denote the available exchanged features, poses, and timestamps by $\mathcal{O}_a$. The local sensor input $x_e$, the victim parameters, internal responses, gradients, and prediction feedback are not part of $\mathcal{O}_a$. In this setting, zero-query means that the attacker does not submit queries to the victim model and does not receive victim predictions for attack updates.

\subsection{Threat Model and Evaluation Objectives}
\label{sec:threat_model}

The ego agent $e$ is the victim and $a\in\mathcal{N}_e$ is a malicious collaborator that modifies its outgoing feature before fusion. The sets $\mathcal{A}_e$ and $\mathcal{B}_e$ collect malicious and benign collaborators, respectively. These labels are used to define evaluation cases and are unavailable to the defense during inference.

\paragraph{White-box reference}
PB is defined under a white-box setting in which the attacker can inspect the victim model $\mathcal{M}_e$, including its architecture, parameters, internal responses, and gradients~\cite{pb}. Its available knowledge is summarized as
\begin{equation}
\begin{aligned}
\mathcal{K}_a^{\mathrm{WB}} =\{\mathcal{O}_a,\operatorname{Arch}(\mathcal{M}_e),\Theta_e, \nabla\mathcal{L}_e\},
\end{aligned}
\label{eq:wb_knowledge}
\end{equation}
where $\Theta_e$ denotes the victim parameters and $\nabla\mathcal{L}_e$ denotes gradients available to the white-box attack.

\paragraph{Zero-query transfer setting}
The transferable attack studied later constructs perturbations using the protocol-observable set $\mathcal{O}_a$ and an attacker-controlled surrogate set $\mathcal{S}=\{\mathcal{M}_s\}_{s=1}^{S}$ ~\cite{transferability}. The attacker has no access to the victim architecture $\operatorname{Arch}(\mathcal{M}_e)$, parameters $\Theta_e$, internal responses, or gradients, and cannot query the victim for prediction feedback. The victim configuration is excluded from $\mathcal{S}$, ensuring that the transferred perturbation is evaluated on an unseen victim.

In both settings, the attacker changes only its transmitted intermediate feature:
\begin{equation}
\begin{aligned}
\tilde{f}_a=f_a+\delta_a, \qquad \delta_a\in\mathcal{C}_{\mathrm{PB}},
\qquad \|\delta_a\|_p\leq\epsilon,
\end{aligned}
\label{eq:attacked_feature}
\end{equation}
where $\epsilon$ is the perturbation budget and $\mathcal{C}_{\mathrm{PB}}$ is the region-aware feasible set defined in Sec.~\ref{sec:pb}. The attacked representation and prediction are
\begin{equation}
\begin{aligned}
\tilde{z}_e
&=\mathcal{F}_e\!\left(f_e,\tilde{f}_a,\{f_j\}_{j\in\mathcal{N}_e\setminus\{a\}}\right),\\
\tilde{Y}_e&=\mathcal{P}_e(\tilde{z}_e).
\end{aligned}
\label{eq:attacked_output}
\end{equation}

At evaluation time, the defense assigns an acceptance decision $g_j\in\{0,1\}$ to each received collaborator. The accepted set and the defended prediction are
\begin{equation}
\begin{aligned}
\mathcal{N}_e^{\mathrm{acc}}&=\{j\in\mathcal{N}_e\mid g_j=1\},\\
z_e^{D}&=\mathcal{F}_e\!\left(f_e,\{\tilde{f}_j\}_{j\in\mathcal{N}_e^{\mathrm{acc}}}\right),\\
Y_e^{D}&=\mathcal{P}_e(z_e^{D}).
\end{aligned}
\label{eq:defended_output}
\end{equation}
For a benign collaborator, $\tilde{f}_j=f_j$, whereas a malicious collaborator may send $\tilde{f}_j=f_j+\delta_j$. 

\begin{figure*}[t]
  \centering
  \includegraphics[width=\linewidth]{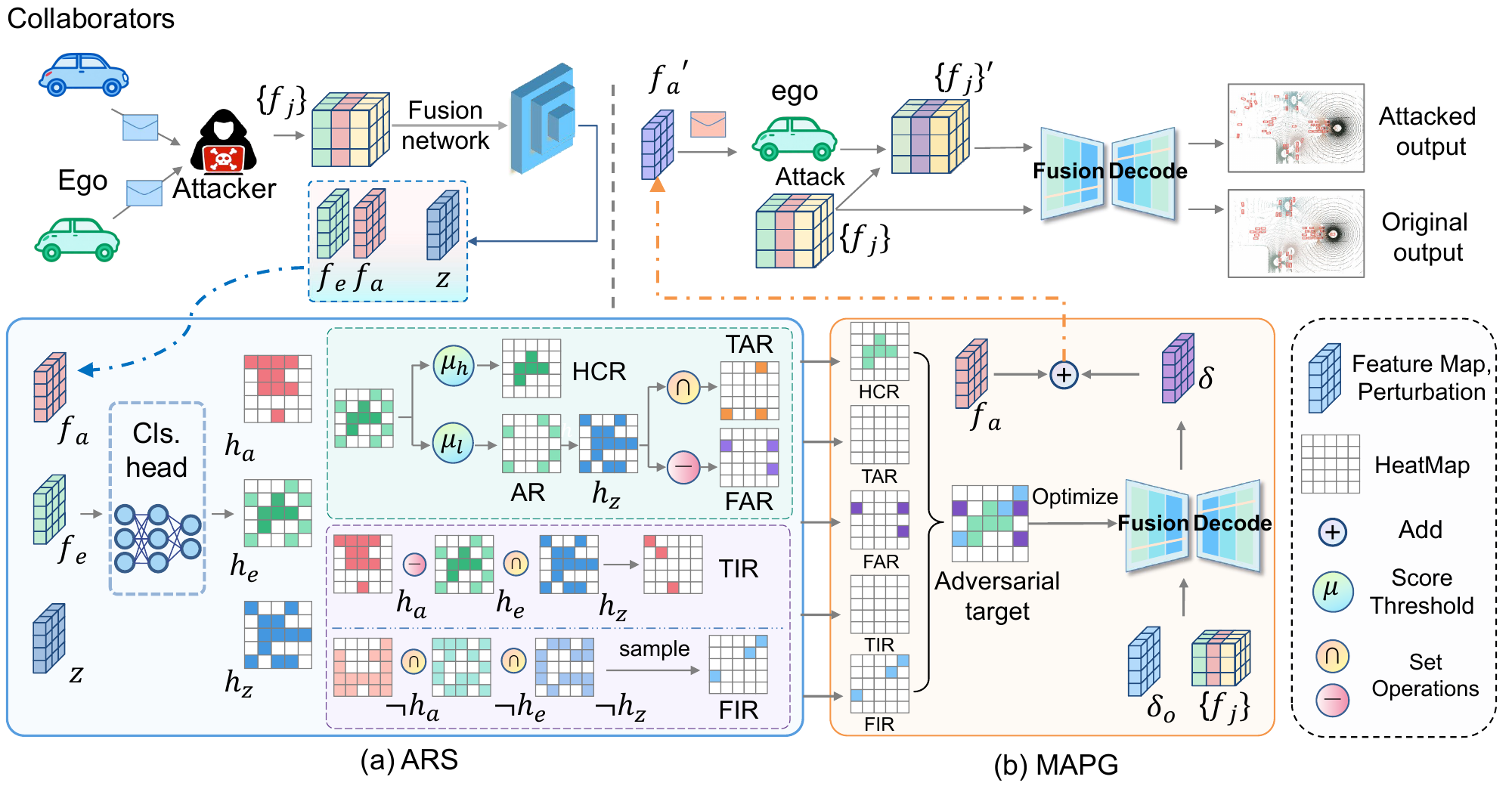}
  \caption{The framework of Pretend Benign (PB). Attack Region Selection (ARS) partitions the BEV perception space into semantically distinct attack regions. Multi-target Adversarial Perturbation Generation (MAPG) then assigns region-specific objectives according to their respective roles in mimicking benign collaboration while degrading the victim's perception.}
  \label{fig:pb}
\end{figure*}

The evaluation follows two complementary perspectives. From the attacker's perspective, we measure the degradation of the victim's perception by comparing the clean prediction $Y_e$ with the attacked prediction $\tilde{Y}_e$. We also report the defense-bypass rate (DBR), defined as the fraction of attacked instances for which the malicious collaborator is accepted, that is, $g_a=1$. From the defender's perspective, we measure the ability to reject malicious collaborators, the ability to retain useful information from benign collaborators, and the recovery of perception after filtering. The first two properties are reflected by the acceptance decisions $g_a$ for $a\in\mathcal{A}_e$ and $g_b$ for $b\in\mathcal{B}_e$, while the recovery property is evaluated from the defended prediction $Y_e^{D}$ relative to the clean cooperative prediction. The experiments use these attack-side and defense-side quantities as the primary evaluation objectives. 
\section{PB: Pretend Benign}
\label{sec:pb}

Pretend Benign (PB)~\cite{pb} is a stealthy white-box attack that exploits the verification asymmetry of cooperative perception to disguise malicious manipulation as benign collaboration. Rather than indiscriminately corrupting transmitted features, PB preserves agreement with the ego agent in reliable regions while manipulating uncertain or imperceptible regions that are difficult to independently verify. As illustrated in Fig.~\ref{fig:pb}, PB consists of Attack Region Selection (ARS), which identifies regions with different semantic roles, and Multi-target Adversarial Perturbation Generation (MAPG), which assigns region-specific objectives to jointly achieve attack effectiveness and stealthiness.

\subsection{Attack Region Selection}

ARS partitions the BEV perception space into semantically distinct regions based on ego confidence, fusion support, and cross-agent visibility, providing region-level semantics for subsequent region-specific attack optimization.

Using the confidence responses and foreground supports defined in Sec.~\ref{sec:pre}, ARS reuses $h_e$, $h_a$, $h_z$ and $\Omega_e^{+}$, $\Omega_a^{+}$, $\Omega_z^{+}$ to assign region semantics under the PB white-box setting. To preserve reliable consensus while exposing uncertain perceptions for manipulation, ARS partitions ego responses using two thresholds:
\begin{equation}
\begin{aligned}
\mathrm{HCR}&=\{u\in\Omega_e^{+}\mid h_e(u)>\mu_h\},\\
\mathrm{AR}&=\{u\in\Omega_e^{+}\mid h_e(u)<\mu_l\},
\end{aligned}
\end{equation}
where $\mu_h>\mu_l$ and $u$ denotes a BEV location. The High-Confidence Region (HCR) contains reliable ego perceptions and is retained to maintain partial consensus, whereas the Ambiguous Region (AR) contains low-confidence perceptions whose correctness cannot be determined from the ego response alone.

\paragraph{Fusion-guided Ambiguity Disambiguation}
Low-confidence evidence alone cannot distinguish true uncertain targets from erroneous predictions. ARS therefore uses the fused response $h_z$, which incorporates complementary observations from multiple agents, as a collaborative pseudo-label:
\begin{equation}
\mathrm{TAR}=\mathrm{AR}\cap\Omega_z^{+},\qquad \mathrm{FAR}=\mathrm{AR}\setminus\Omega_z^{+}.
\end{equation}
The True Ambiguous Region (TAR) contains low-confidence ego perceptions supported by collaborative fusion and is thus more likely to correspond to real objects, whereas the False Ambiguous Region (FAR) contains unsupported predictions that are more likely erroneous. This distinction enables PB to apply opposite attack behaviors to true and false ambiguous perceptions.

\paragraph{Imperceptible Region Modeling}
Cooperative perception is particularly valuable in regions invisible to the ego but observable from another viewpoint. To identify such useful supplementary information, ARS selects attacker-visible but ego-imperceptible regions that are also supported by fusion:
\begin{equation}
\mathrm{TIR}=(\Omega_a^{+}\setminus\Omega_e^{+})\cap\Omega_z^{+}.
\end{equation}
The True Imperceptible Region (TIR) denotes regions that are more likely to correspond to objects detected by the attacker but missed by the ego, representing complementary information that cooperative perception is intended to provide.

PB further exploits background regions to fabricate plausible supplementary information. Candidate background locations are defined as
\begin{equation}
\Omega_{\mathrm{bg}}=(\Omega\setminus\Omega_a^{+})\cap(\Omega\setminus\Omega_e^{+})\cap(\Omega\setminus\Omega_z^{+}),
\end{equation}
where $\Omega$ denotes the complete BEV perception space. Arbitrary background sampling could produce implausible false positives and weaken stealthiness. PB therefore favors locations closer to the attacker and farther from the ego through pose-aware sampling:
\begin{equation}
\mathrm{FIR}=\Psi_{\mathrm{pose}}\!\left(\Omega_{\mathrm{bg}},p_a,p_e\right),
\end{equation}
where $p_a$ and $p_e$ denote the attacker and ego poses. The resulting False Imperceptible Region (FIR) provides plausible locations for fabricated supplementary targets.

ARS thus produces five working regions. HCR preserves reliable consensus, TAR and TIR encode putative supported evidence, and FAR and FIR encode putative unsupported or fabricated evidence. These labels reflect support relations inferred from model responses rather than ground-truth object status, forming the basis for the region-specific attack objectives in MAPG.

\subsection{Multi-target Adversarial Perturbation Generation}

Given the ARS regions, Multi-target Adversarial Perturbation Generation (MAPG) assigns different attack behaviors according to their roles in collaboration. HCR preserves sufficient consensus, TAR and TIR suppress likely true information, while FAR and FIR reinforce or fabricate false information. This allows PB to imitate benign collaborative behavior while degrading the victim's perception.

\paragraph{Region-specific Attack Objectives}
Starting from an initial perturbation $\delta_a^{(0)}$, MAPG iteratively optimizes $\delta_a$ through the differentiable victim pipeline. For region $r$, let $y_r$ denote its reference detection when available, and let $\tilde{y}_r$ and $\tilde{s}_r$ denote the corresponding adversarial detection and confidence score. The region-specific loss is
\begin{equation}
\mathcal{L}_{\mathrm{adv}}^{r}=
\begin{cases}
\begin{aligned}
&-\log(1-\tilde{s}_r)\mathrm{IoU}(\tilde{y}_r,y_r)\\
&\quad+\lambda_c\mathcal{L}_{\mathrm{cons}}(\tilde{y}_r,y_r),
\end{aligned}
& r\in\mathrm{HCR},\\
-\log(1-\tilde{s}_r)\mathrm{IoU}(\tilde{y}_r,y_r),
& r\in\{\mathrm{TAR},\mathrm{TIR}\},\\
-\log(\tilde{s}_r),
& r\in\{\mathrm{FAR},\mathrm{FIR}\}.
\end{cases}
\end{equation}

TAR and TIR are suppressed to remove likely true ego or supplementary information, while FAR and FIR are strengthened to confirm erroneous perceptions or fabricate plausible objects. These complementary objectives mimic the correction, confirmation, and supplementation functions of benign collaboration while producing the opposite effect on perception.

\paragraph{Consensus-preserving Regularization}
Excessive degradation of HCR would destroy the reliable consensus required for disguise. PB therefore constrains HCR confidence and localization quality through
\begin{equation}
\begin{aligned}
\mathcal{L}_{\mathrm{cons}}(\tilde{y}_r,y_r)
=&-\log\left(1-\sqrt{[\tau_{\mathrm{score}}-\tilde{s}_r]_{+}}\right)\\
&-\log\left(1-\sqrt{[\tau_{\mathrm{IoU}}^{\mathrm{PB}}-\mathrm{IoU}(\tilde{y}_r,y_r)]_{+}}\right),
\end{aligned}
\end{equation}
where $[x]_{+}=\max(x,0)$, and $\tau_{\mathrm{score}}$ and $\tau_{\mathrm{IoU}}^{\mathrm{PB}}$ denote the minimum confidence and IoU required for reliable consensus. The positive-part operator keeps the square-root arguments nonnegative and makes the penalty vanish once the corresponding lower bound is satisfied.

\paragraph{Perturbation Optimization}
The region-wise objectives are jointly optimized under the perturbation budget defined in Sec.~\ref{sec:threat_model}:
\begin{equation}
\delta_a^{*}=\arg\min_{\|\delta_a\|_p\leq\epsilon}\sum_{r}\mathcal{L}_{\mathrm{adv}}^{r}.
\end{equation}
The optimized perturbation is injected as $\tilde{f}_a=f_a+\delta_a^{*}$ and transmitted for cooperative fusion. By preserving reliable consensus while suppressing true information and enhancing or fabricating false information, PB disguises malicious manipulation as seemingly benign collaboration.
\section{Zero-Query Black-box Pretend Benign Attacks}
\label{sec:pb++}

\begin{figure*}[t]
  \centering
  \includegraphics[width=\linewidth]{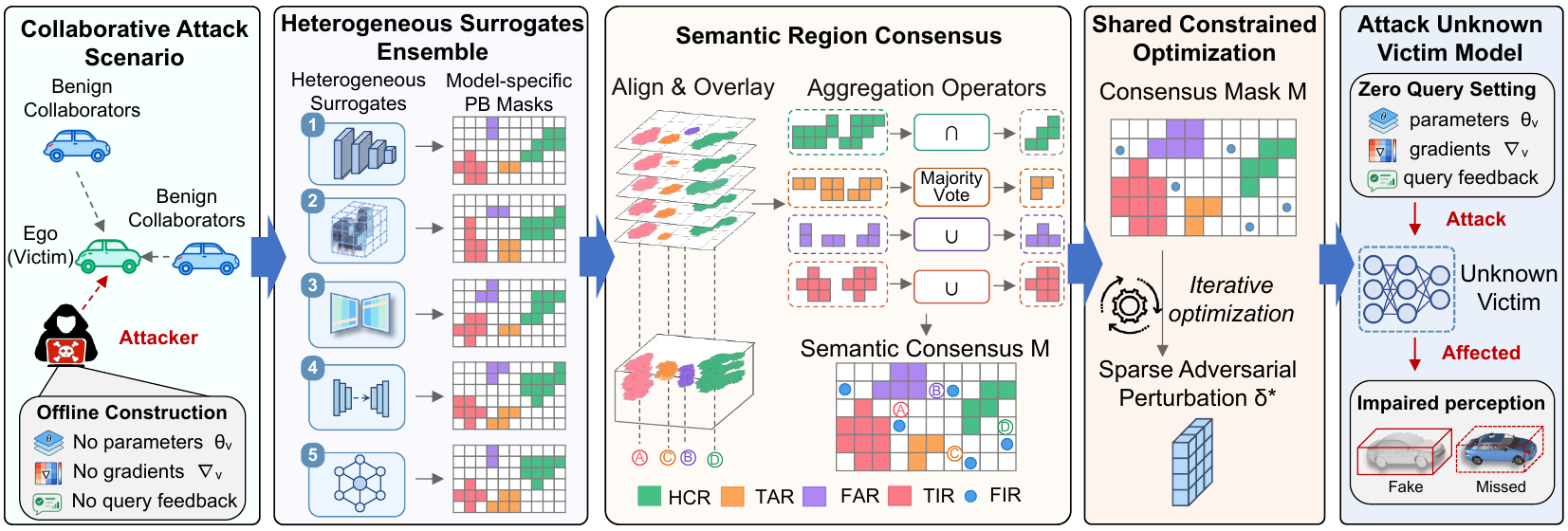}
  \caption{Offline construction of transferable PB knowledge with TrPB. Each heterogeneous surrogate generates PB-region responses for the same cooperative scene. Semantic Region Consensus (SRC) aggregates these responses by region role, and Shared Constrained Optimization (SCO) learns a shared perturbation under the PB support and magnitude constraints. The resulting consensus masks and perturbation patterns are exported to RT-TrPB for online reuse.}
  \label{fig:TrPB}
\end{figure*}

PB achieves stealthy manipulation by exploiting the verification asymmetry of cooperative perception, but its white-box realization relies on victim-specific responses and gradients. Such privileged access is difficult to assume in practical cooperative environments, where heterogeneous agents may deploy private, unseen perception models and victim feedback is unavailable for iterative querying. We therefore extend PB along one offline-to-online attack path with complementary roles. TrPB constructs transferable PB knowledge offline from heterogeneous surrogate models, whereas RT-TrPB freezes and reuses that knowledge to generate each online perturbation through a single synthesis path. Only RT-TrPB is executed at attack time. 

\subsection{TrPB: Transferable Pretend Benign}
\label{sec:trpb}

Transferable Pretend Benign (TrPB) is the offline construction branch of the zero-query attack path. It uses an attacker-controlled surrogate set to convert PB's region-specific stealth objective into two reusable knowledge objects: consensus region semantics for deciding where to attack and a shared constrained perturbation for deciding how to attack. Directly reusing the region masks and perturbation optimized on a single surrogate can overfit its confidence responses, feature geometry, and fusion behavior. As a result, the masks may encode surrogate-specific support relations, while the perturbation direction may be poorly aligned with an unseen victim's representation and fusion rule~\cite{transferability, black-transfer-1}. As illustrated in Fig.~\ref{fig:TrPB}, TrPB addresses these sources of overfitting through Heterogeneous Surrogate Ensemble (HSE), Semantic Region Consensus (SRC), and Shared Constrained Optimization (SCO), jointly extracting transferable attack regions and perturbation behaviors that determine \emph{where to attack} and \emph{how to attack}.

\paragraph{Heterogeneous Surrogate Ensemble}
To enable PB perturbations to transfer to unseen victim models under the zero-query black-box setting, HSE constructs an attacker-side surrogate set $\mathcal{S}=\{\mathcal{M}_s\}_{s=1}^{S}$ from diverse and representative cooperative perception models with heterogeneous perception backbones and fusion strategies. For the same cooperative scene, different surrogates provide complementary confidence, fusion, and vulnerability responses, allowing model-specific behavior to be distinguished from attack characteristics that remain stable across architectures~\cite{black-transfer-2, black-transfer-3}. This ensemble view provides the construction basis for the region consensus and shared optimization defined below. It does not by itself guarantee transfer to every victim family.

\paragraph{Semantic Region Consensus}
Heterogeneous models inevitably disagree on PB attack regions. Uniform intersection would discard excessive attack opportunities, whereas uniform union could retain unstable model-specific regions and weaken stealthiness. SRC therefore first performs PB region selection independently on each surrogate using model-specific, independently calibrated confidence thresholds, and then aggregates the resulting regions according to their semantic roles:
\begin{equation}
\begin{aligned}
\widehat{\mathrm{HCR}} &= \bigcap_{s=1}^{S}\mathrm{HCR}^{(s)}, & \widehat{\mathrm{TAR}} &= \operatorname{Vote}\!\left(\{\mathrm{TAR}^{(s)}\}_{s=1}^{S}\right),\\
\widehat{\mathrm{FAR}} &= \bigcup_{s=1}^{S}\mathrm{FAR}^{(s)}, & \widehat{\mathrm{TIR}} &= \bigcup_{s=1}^{S}\mathrm{TIR}^{(s)}.
\end{aligned}
\end{equation}
HCR uses intersection to retain reliable cross-model consensus, TAR uses strict majority voting to preserve stable uncertain targets, while FAR and TIR use union to maintain sufficient attack coverage. Since FIR is sampled from background rather than inferred from existing detections, it is resampled after the above regions are aggregated using the pose-aware strategy in Sec.~\ref{sec:pb}. SRC therefore provides a role-aware approximation of \emph{where to attack} while retaining the distinct PB functions of consensus preservation, supported-region suppression, and unsupported-region reinforcement.

\paragraph{Shared Constrained Optimization}
Consensus regions alone do not guarantee transferable perturbations, since adversarial directions optimized for an individual surrogate may remain model-specific. SCO therefore optimizes a single shared perturbation across the surrogate ensemble through compatible BEV feature interfaces:
\begin{equation}
\begin{aligned}
\delta_a^{\mathrm{Tr}*}
=\arg\min_{\delta_a\in\mathcal{C}_{\mathrm{PB}}}
\Bigg[
\frac{1}{S}\sum_{s=1}^{S}\mathcal{L}_{\mathrm{PB}}^{(s)}(\delta_a)
&\\
+\lambda_{\mathrm{mom}}\mathcal{L}_{\mathrm{mom}}
+\lambda_{\mathrm{norm}}\mathcal{L}_{\mathrm{norm}}
&\Bigg].
\end{aligned}
\end{equation}
where $\mathcal{L}_{\mathrm{PB}}^{(s)}$ is the PB objective evaluated on surrogate $\mathcal{M}_s$, $\mathcal{L}_{\mathrm{mom}}$ reduces discrepancies in perturbation-induced feature statistics across surrogates, and $\mathcal{L}_{\mathrm{norm}}$ penalizes excessive perturbation magnitude. The feasible set $\mathcal{C}_{\mathrm{PB}}$ restricts perturbations to the five consensus PB regions fixed before optimization, with each update projected onto the corresponding region mask and clipped to its region-specific magnitude bounds. Consequently, SCO transfers not only adversarial effectiveness but also the benign-like collaborative behavior required for stealthiness.

TrPB derives consensus regions and perturbations solely from the surrogate ensemble, enabling zero-query PB attacks on unseen victim models by transferring region semantics and perturbation behaviors without victim-specific knowledge or feedback. However, generating these outputs for each new sample still requires multi-surrogate inference, SRC aggregation, and SCO optimization. TrPB therefore exports its consensus masks and shared perturbation patterns to RT-TrPB, where they are converted into a frozen region predictor and a reusable perturbation dictionary for online synthesis.

\begin{figure*}[t]
  \centering
  \includegraphics[width=\linewidth]{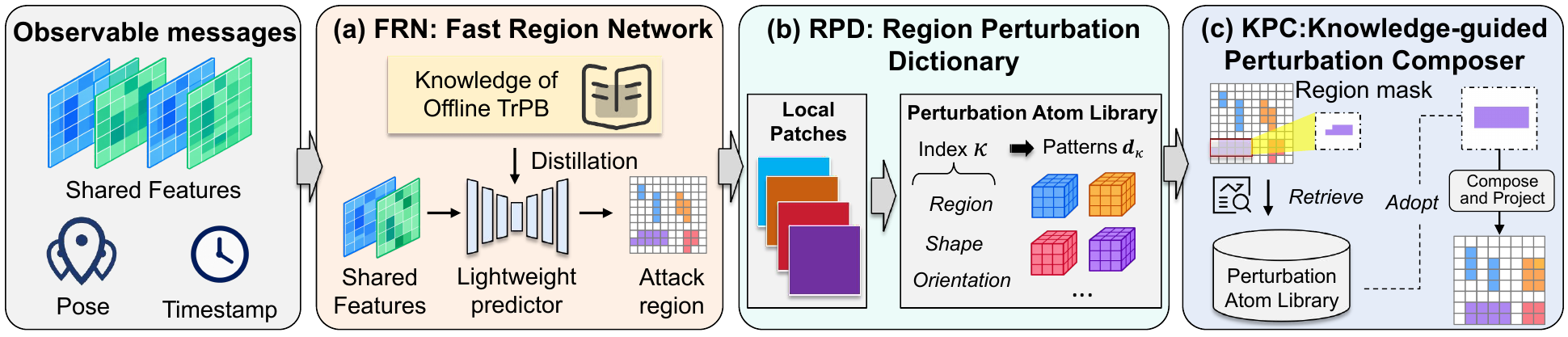}
  \caption{Online realization (RT-TrPB) of the transferable PB knowledge produced by TrPB. (a) The frozen Fast Region Network (FRN) predicts the four inferred PB regions from protocol-observable features, (b) the frozen Region Perturbation Dictionary (RPD) stores region-keyed perturbation atoms derived from the TrPB perturbations. (c) Knowledge-guided Perturbation Composer (KPC) retrieves, adapts, and projects compatible atoms to generate online attack perturbations. Only this synthesis path is executed online, without surrogate inference, iterative optimization, or victim prediction queries.}
  \label{fig:RT-TrPB}
\end{figure*}

\subsection{Real-Time Transferable Pretend Benign}
\label{sec:rt_trpb}

As illustrated in Fig.~\ref{fig:RT-TrPB}, RT-TrPB realizes the two TrPB knowledge objects by replacing per-sample HSE--SRC computation with an offline-trained Fast Region Network (FRN), and per-sample SCO optimization with a Region Perturbation Dictionary (RPD) and a Knowledge-guided Perturbation Composer (KPC). The FRN distills SRC's knowledge of \emph{where to attack} into transferable region semantics, while the RPD encodes SCO's knowledge of \emph{how to perturb} as reusable perturbation patterns~\cite{distilling, CrossKD}. The FRN and RPD are frozen before deployment, and KPC combines their outputs to synthesize sample-specific attacks online.

\paragraph{Fast Region Network}
Repeating multi-surrogate region selection for every incoming frame
would prevent efficient online deployment. FRN therefore distills
the consensus regions generated by offline TrPB into a lightweight
predictor. Under the evaluated message protocol, let $\mathcal{I}_a$
denote the set of collaborator features available to the attacker,
whose mean is computed as
\begin{equation}
\bar{f}_a=\frac{1}{|\mathcal{I}_a|}
\sum_{i\in\mathcal{I}_a}f_i.
\end{equation}
For an attacker-only configuration, $\mathcal{I}_a=\{a\}$,
yielding $\bar{f}_a=f_a$. Using the ego feature $f_e$, observable
through normal feature sharing, and the aggregated attacker feature
$\bar{f}_a$, we construct a compact 12-channel input for FRN:
\begin{equation}
\begin{aligned}
\mathbf{x}_{\mathrm{FRN}}
&=\left[\phi(f_e),\phi(\bar{f}_a),
\mu_C(\Delta f_a),\mu_C(|\Delta f_a|),\xi_y,\xi_x\right],\\
\phi(f)
&=\left[\mu_C(f),\sigma_C(f),
\mu_C(|f|),\sqrt{\mu_C(f^2)}\right],
\end{aligned}
\end{equation}
where $\Delta f_a=\bar{f}_a-f_e$, $\mu_C$ and $\sigma_C$
denote cross-channel mean and standard deviation, and
$(\xi_y,\xi_x)$ are normalized BEV coordinates. Each statistical
channel is independently normalized.

FRN predicts four region logits, which are independently decoded by selecting the top-$K_r$ spatial locations:
\begin{equation}
\begin{aligned}
\boldsymbol{\ell}
&=\mathcal{G}_{\mathrm{FRN}}(\mathbf{x}_{\mathrm{FRN}}),\\
\hat{m}_r(u)
&=\mathbb{I}\left[
u\in\operatorname{TopK}_{K_r}(\ell_r)
\right],
\quad r\in\mathcal{R},
\end{aligned}
\end{equation}
where $\mathcal{R}=\{\mathrm{HCR},\mathrm{TAR},\mathrm{FAR},\mathrm{TIR}\}$ and $\ell_r$ denotes the corresponding region logits. During offline training, the SRC masks from TrPB supervise the four logits with weighted binary cross-entropy and Soft Dice losses. At deployment, FRN predicts only the four inferred regions in $\mathcal R$, while FIR is sampled separately as part of the complete PB region set
$\mathcal R_{\mathrm{PB}}$. FRN therefore amortizes the \emph{where to attack} output of TrPB rather than redefining the PB region objective.

\paragraph{Region Perturbation Dictionary}
Region prediction alone still leaves the iterative perturbation optimization of SCO. Although SCO produces scene-specific perturbations, local patches serving the same region-specific role tend to exhibit similar patterns across different scenes. RPD exploits this cross-scene regularity by extracting and storing these local patches as reusable primitives indexed by $\kappa=(r,b_{\mathrm{shp}},b_{\mathrm{ori}})$, where $r\in\mathcal R_{\mathrm{PB}}$ denotes the PB region type and the remaining fields characterize the support shape and orientation. The dictionary thus captures reusable local perturbation patterns according to their region-specific roles, approximating SCO's \emph{how to perturb} output without introducing a new unconstrained perturbation objective. For each key, candidate patches are aligned to a representative geometry and consolidated into a prototype atom:
\begin{equation}
\mathbf{d}_{\kappa}
=\operatorname{Clip}_{\epsilon_r}
\left(
\operatorname{sgn}\!\left(
\sum_{i=1}^{N_\kappa}\operatorname{sgn}(\bar{\delta}_i)
\right)
\odot
\frac{1}{N_\kappa}\sum_{i=1}^{N_\kappa}|\bar{\delta}_i|
\right),
\end{equation}
where $\bar{\delta}_i$ denotes an aligned candidate patch, $N_\kappa$ is the number of candidates sharing key $\kappa$, and $\epsilon_r$ is the corresponding region-wise perturbation bound. This sign-and-magnitude aggregation preserves recurrent perturbation directions and strengths while suppressing sample-specific variations. RPD thereby converts the optimized knowledge of \emph{how to perturb} into compact patterns reusable across scenes. 

\paragraph{Knowledge-guided Perturbation Composer}
At runtime, KPC combines the regions predicted by FRN with the perturbation atoms retrieved from RPD. For each decoded connected region $q$, it constructs a region--shape--orientation key and retrieves the corresponding atom. If an exact match is unavailable, retrieval progressively relaxes to the same region and orientation, then to the same region type, selecting the closest-sized atom. The retrieved atom is resized, masked to the target support, and placed at the predicted BEV location. The resulting perturbation is
\begin{equation}
\delta_a^{\mathrm{RT}}=\Pi_{\mathrm{PB}}\left(\sum_{q\in\mathcal{Q}}\mathcal{T}_q(\mathbf{d}_{\kappa_q})\right),
\end{equation}
where $\Pi_{\mathrm{PB}}$ denotes projection onto the PB feasible set $\mathcal C_{\mathrm{PB}}$, $\mathcal{Q}$ denotes the decoded attack regions, $\kappa_q$ is the retrieved dictionary key, and $\mathcal{T}_q(\cdot)$ performs geometric adaptation and placement. The projection $\Pi_{\mathrm{PB}}(\cdot)$ suppresses perturbations outside the selected PB regions and enforces their respective magnitude bounds, preserving the region-specific attack semantics. The reuse path is detailed in Fig.~\ref{fig:RT-TrPB} (c). Finally, $\delta_a^{\mathrm{RT}}$ is directly injected into the current feature as $\tilde{f}_a=f_a+\delta_a^{\mathrm{RT}}$ and transmitted for cooperative fusion. 

Since all RT-TrPB components are learned and frozen offline, the online stage requires neither surrogate-model inference nor gradient-based refinement and remains independent of victim predictions or feedback. Thus, TrPB supplies transferable region and perturbation knowledge offline, and RT-TrPB reuses these frozen objects to produce attack perturbations online without surrogate inference, iterative optimization, or victim prediction feedback. The deployment path therefore executes RT-TrPB alone, while TrPB remains the offline source of its supervision and dictionary entries.
\begin{figure}[t]
  \centering
   \includegraphics[width=\linewidth]{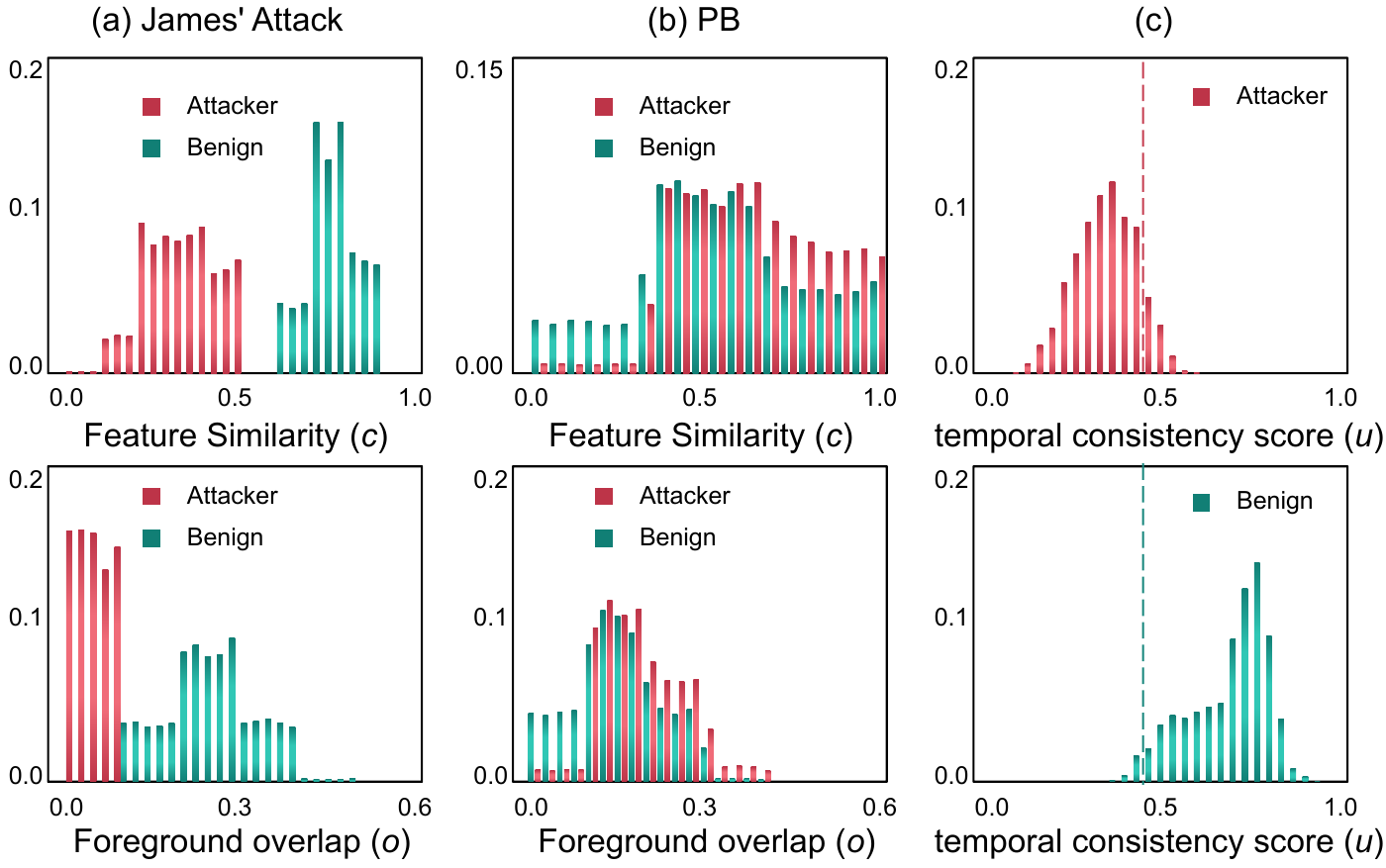}
   \caption{Empirical motivation for temporal verification. Panels (a) and (b) compare feature similarity and foreground overlap for James' Attack~\cite{James_attack} and PB~\cite{pb} against benign collaborators. Panel (c) shows the temporal consistency score distributions. The evaluated PB traces retain local spatial agreement yet exhibit a separable temporal pattern, motivating the motion-aware and historical evidence used by STTrust.}
   \label{fig:temporal_spatial_analyse}
\end{figure}

\begin{figure*}[t]
  \centering
  \includegraphics[width=\linewidth]{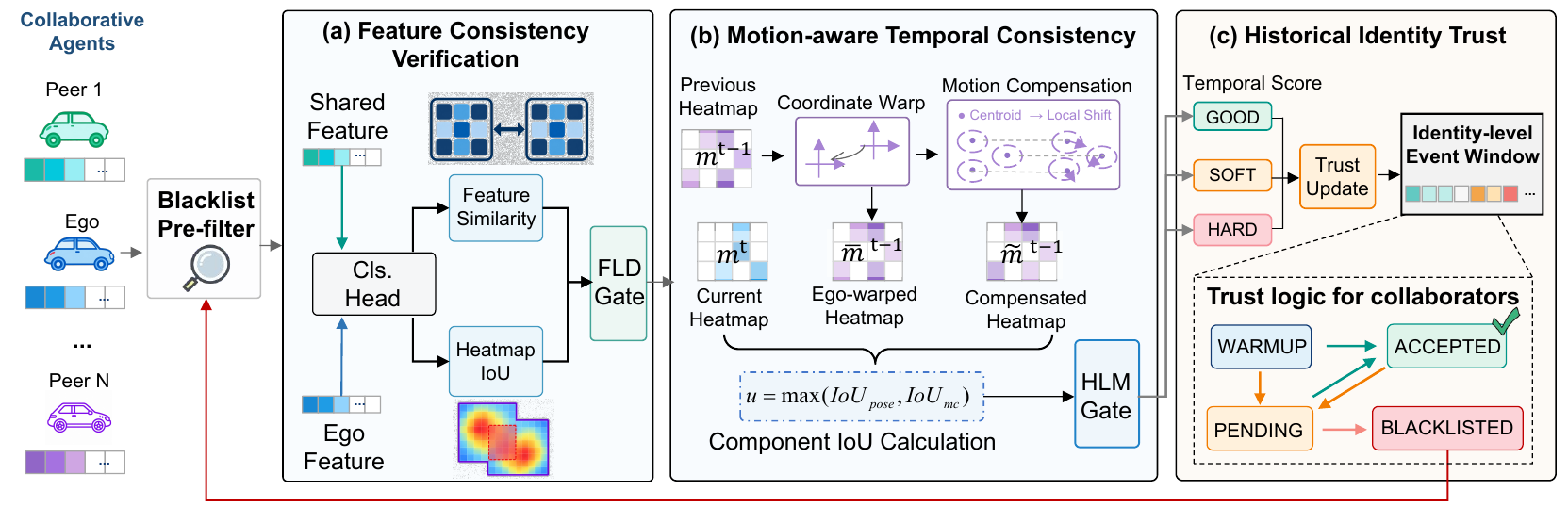}
  \caption{The framework of the Spatio-Temporal Trust Framework (STTrust). STTrust progressively verifies collaborative agents through (a) Feature Consistency Verification (FCV), which filters obvious current-frame inconsistencies using feature and spatial consistency, (b) Motion-aware Temporal Consistency (MTC), which evaluates temporal coherence after ego-motion alignment and target-motion compensation, and (c) Historical Identity Trust (HIT), which accumulates graded temporal evidence to maintain identity-level trust states and reject persistently suspicious collaborators.}
  \label{fig:STTrust}
\end{figure*}

\section{Spatio-Temporal Trust Defense}
\label{sec:sttrust}

Existing collaborative perception defenses use feature-level and spatial consistency as current-frame evidence~\cite{pb,made,LatentGuardian,GLST}. However, PB and its black-box extensions challenge consistency-based defenses by preserving benign-like behavior in locally verifiable regions. As shown by the empirical distributions in Fig.~\ref{fig:temporal_spatial_analyse}, evaluated PB messages can retain local spatial agreement, whereas temporal consistency scores provide additional evidence for distinguishing malicious from benign collaborators. Motivated by these findings, we develop the \emph{Spatio-Temporal Trust Framework} (STTrust), which integrates complementary spatial, temporal, and historical evidence to verify collaborators. As illustrated in Fig.~\ref{fig:STTrust}, STTrust comprises Feature Consistency Verification (FCV), Motion-aware Temporal Consistency (MTC), and Historical Identity Trust (HIT), and operates as a lightweight plug-in before cooperative fusion without modifying the original perception model.

\paragraph{Feature Consistency Verification}
Stealthy attacks may preserve partial agreement with the ego agent, but strongly inconsistent information can still be efficiently filtered within the current frame. For collaborator $j$ at frame $t$, we obtain its confidence response and foreground mask as
\begin{equation}
h_j^t=\mathcal{P}_{\mathrm{cls}}(\tilde{f}_j^t),\qquad m_j^t=\mathbf{1}[h_j^t\geq\tau_f],
\end{equation}
with analogous definitions for the ego agent. FCV jointly evaluates feature similarity within the shared salient region and the spatial overlap between collaborator and ego responses:
\begin{equation}
c_j^t=\operatorname{Sim}_{m_j^t\cap m_e^t}(\tilde{f}_j^t,f_e^t),\qquad o_j^t=\operatorname{IoU}(m_j^t,m_e^t).
\end{equation}
Here, $\operatorname{Sim}$ averages the cosine similarity of feature vectors over spatial locations shared by both foreground masks. The FCV decision is defined as $v_j^t=\mathbf{1}[c_j^t>\tau_{\mathrm{sim}}\land o_j^t>\tau_{\mathrm{IoU}}]$, where both criteria must be satisfied for acceptance. Collaborators failing FCV are rejected from current-frame fusion and contribute \textit{HARD} evidence to historical trust assessment, while those passing proceed to temporal verification. FCV thus serves as an efficient risk filter rather than the final trust decision.

\paragraph{Motion-aware Temporal Consistency}
Single-frame consistency is insufficient against PB-style attacks, as malicious collaborators can craft messages that resemble benign behavior locally, whereas genuine collaborators should exhibit perception responses that evolve consistently with the physical scene. Although temporal and historical cues have been explored in prior collaborative perception defenses~\cite{GCP,PRBI}, MTC operationalizes temporal consistency through explicit motion compensation and foreground-mask comparison. Since ego motion and dynamic targets make direct comparison of consecutive BEV responses unreliable, MTC first warps the previous response into the current ego coordinate system:
\begin{equation}
\bar{m}_j^{t-1}=\mathcal{W}\left(m_j^{t-1},(T_e^t)^{-1}T_e^{t-1}\right),
\end{equation}
and then compensates for target motion through lightweight foreground-component matching before comparing the resulting masks using IoU. Specifically, foreground components are extracted from the aligned historical and current masks and greedily matched based on their centroid distances. Matched historical components are then shifted according to their centroid displacements, yielding the motion-compensated mask $\widetilde{m}_j^{t-1}$. To avoid penalizing benign collaborators when motion matching is uncertain, the temporal consistency score is defined as
\begin{equation}
u_j^t=\max\left\{\operatorname{IoU}(m_j^t,\bar{m}_j^{t-1}),\operatorname{IoU}(m_j^t,\widetilde{m}_j^{t-1})\right\}.
\end{equation}
Rather than making a brittle binary decision, MTC maps valid temporal scores into three evidence levels:
\begin{equation}
\eta_j^t=
\begin{cases}
\mathit{H}, & u_j^t\leq\tau_{\mathrm{hard}},\\
\mathit{S}, & \tau_{\mathrm{hard}}<u_j^t<\tau_{\mathrm{soft}},\\
\mathit{G}, & u_j^t\geq\tau_{\mathrm{soft}},
\end{cases}
\end{equation}
where $\mathit{H}$, $\mathit{S}$, and $\mathit{G}$ denote \textit{HARD}, \textit{SOFT}, and \textit{GOOD} evidence, respectively. This graded representation separates strong temporal anomalies from moderate deviations caused by normal perception uncertainty. When reliable temporal comparison is unavailable for a newly observed
collaborator, a temporally discontinuous frame, or an invalid ego pose, MTC treats the temporal evidence as unavailable, filters out the current message, and skips historical trust updating.

\paragraph{Historical Identity Trust}
Temporal evidence from a single frame can be noisy: benign collaborators may occasionally deviate, while stealthy attackers may produce repeated mild anomalies. Following prior work on historical and identity-level trust in collaborative sensing~\cite{PRBI,Hallyburton2025MATE}, HIT accumulates evidence for each stable collaborator identity and maintains a continuous trust score $\rho_j^t$ and a discrete identity state
$\sigma_j^t\in\{\mathit{WU},\mathit{AC},\mathit{PD},\mathit{BL}\}$, corresponding to \textit{WARMUP}, \textit{ACCEPTED}, \textit{PENDING}, and \textit{BLACKLISTED}, respectively.

New collaborators enter \textit{WU} with initial trust $\rho_{\mathrm{init}}$. Each consecutive \textit{GOOD} observation increases their trust by $1/N_{\mathrm{WU}}$. After $N_{\mathrm{WU}}$ consecutive \textit{GOOD} observations, they enter \textit{AC} if the trust requirement is met. For established identities, trust is updated according to the current evidence:
\begin{equation}
\rho_j^t=
\begin{cases}
\rho_j^{t-1}+\dfrac{1-\rho_j^{t-1}}{R},
& \eta_j^t=\mathit{G},\\
[\rho_j^{t-1}-\lambda_S]_+,
& \eta_j^t=\mathit{S},\\
[\rho_j^{t-1}-\lambda_H]_+,
& \eta_j^t=\mathit{H},
\end{cases}
\end{equation}
where $\mathit{G}$, $\mathit{S}$, and $\mathit{H}$ denote \textit{GOOD}, \textit{SOFT}, and \textit{HARD} evidence, respectively, and $[x]_+=\max(x,0)$ ensures that trust never falls below zero. A \textit{GOOD} observation gradually restores trust toward 1, whereas \textit{SOFT} and \textit{HARD} evidence reduce it, with $\lambda_H>\lambda_S$ imposing a larger penalty for \textit{HARD} evidence.

To distinguish persistent anomalies from occasional deviations, HIT separately accumulates anomaly mass over the latest $W$ evidence events:
\begin{equation}
A_j^t=\sum_{k\in\mathcal{H}_j^t}w(\eta_j^k),
\qquad
w(\mathit{G})=0,\quad
w(\mathit{S})=1,\quad
w(\mathit{H})=2,
\end{equation}
where $\mathcal{H}_j^t$ denotes the recent evidence window. The thresholds $B_{\mathrm{PD}}$ and $B_{\mathrm{BL}}$ determine when an identity enters \textit{PD} and \textit{BL}, respectively, with $B_{\mathrm{BL}}>B_{\mathrm{PD}}$. Reaching $B_{\mathrm{PD}}$ triggers \textit{PD}, while reaching $B_{\mathrm{BL}}$ results in irreversible blacklisting. A \textit{PD} identity returns to \textit{AC} only after $N_{\mathrm{rec}}$ consecutive \textit{GOOD} observations and once its trust score satisfies $\rho_j^t\geq\tau_{\mathrm{trust}}$. Upon recovery, its anomaly history is cleared. \textit{BL} identities are excluded from subsequent verification. The state transitions are illustrated in Fig.~\ref{fig:STTrust} (c).

The final acceptance decision jointly considers current-frame consistency, temporal evidence, and historical trust:
\begin{equation}
g_j^t=\mathbf{1}\!\left[v_j^t=1,\;\eta_j^t=\mathit{G},\;\sigma_j^t=\mathit{AC},\;\rho_j^t\geq\tau_{\mathrm{trust}}\right].
\label{eq:final_accept}
\end{equation}
The conjunction in Eq.~\eqref{eq:final_accept} makes the acceptance rule explicit: a message must pass the current-frame gate, provide GOOD temporal evidence, and come from an identity in the ACCEPTED state with sufficient trust. By progressively integrating spatial-feature consistency, motion-aware temporal coherence, and identity-level historical trust, STTrust advances collaborator verification from instantaneous anomaly detection to continuous evidence accumulation, targeting the temporal anomalies exhibited by the evaluated PB-style attacks while preserving a history-aware acceptance rule.
\section{Experiments}
\label{sec:exp}

\begin{table*}[ht]
\caption{White-box attack performance on OPV2V~\cite{opv2v}, V2XSet~\cite{v2xset}, and V2V4Real~\cite{V2v4real} using PointPillars~\cite{pointpillars} with V2VNet~\cite{v2vnet} as the victim model. AP is evaluated at IoU thresholds of 0.3, 0.5, and 0.7. $\Delta$AP denotes the AP decrease relative to the No Attack baseline, while DBR (Defense-Bypass Rate) measures the percentage of attacks that successfully bypass the corresponding defense.}
\centering
\setlength{\tabcolsep}{4.5pt}
\resizebox{\textwidth}{!}{

\begin{tabular}{ccccccccc}
\hline
\multirow{3}{*}{Dataset}
& \multirow{3}{*}{Attack}
& \multirow{2}{*}{No Defense}
& \multicolumn{3}{c}{\multirow{2}{*}{ROBOSAC Defense~\cite{robosac}}}
& \multicolumn{3}{c}{\multirow{2}{*}{FLD Defense~\cite{pb}}} \\
& & & \multicolumn{3}{c}{} & \multicolumn{3}{c}{} \\
\cmidrule(lr){3-3}\cmidrule(lr){4-6}\cmidrule(lr){7-9}
& & AP@(0.3 / 0.5 / 0.7)
& AP@(0.3 / 0.5 / 0.7) $\downarrow$
& $\Delta$AP@(0.3 / 0.5 / 0.7) $\uparrow$
& DBR $\uparrow$
& AP@(0.3 / 0.5 / 0.7) $\downarrow$
& $\Delta$AP@(0.3 / 0.5 / 0.7) $\uparrow$
& DBR $\uparrow$ \\ \hline 

\multirow{6}{*}{OPV2V~\cite{opv2v}}
& No Attack & 0.91 / 0.91 / 0.85 & - & - & - & - & - & - \\
& PGD~\cite{pgd} & 0.40 / 0.31 / 0.04 & 0.89 / 0.88 / 0.76 & 0.02 / 0.03 / 0.09 & 0.00\% & 0.90 / 0.89 / 0.80 & 0.01 / 0.02 / 0.05 & 0.00\% \\
& C\&W~\cite{cw-l2} & 0.38 / 0.29 / 0.04 & 0.89 / 0.88 / 0.76 & 0.02 / 0.03 / 0.09 & 0.00\% & 0.90 / 0.89 / 0.80 & 0.01 / 0.02 / 0.05 & 0.00\% \\
& James' Attack~\cite{James_attack} & 0.79 / 0.74 / 0.23 & 0.89 / 0.88 / 0.76 & 0.02 / 0.03 / 0.09 & 0.00\% & 0.90 / 0.89 / 0.80 & 0.01 / 0.02 / 0.05 & 0.00\% \\
& AdvGPS~\cite{advgps} & 0.60 / 0.59 / 0.42 & 0.70 / 0.69 / 0.55 & 0.21 / 0.22 / 0.30 & 76.20\% & 0.90 / 0.89 / 0.80 & 0.01 / 0.02 / 0.05 & 0.00\% \\
\rowcolor{gray!12}
& \textbf{PB(Ours)} & 0.54 / 0.53 / 0.42 & \textbf{0.54 / 0.53 / 0.42} & \textbf{0.37 / 0.38 / 0.43} & \textbf{100.00\%} & \textbf{0.54 / 0.53 / 0.42} & \textbf{0.37 / 0.38 / 0.43} & \textbf{100.00\%} \\ \hline

\multirow{6}{*}{V2XSet~\cite{v2xset}}
& No Attack & 0.87 / 0.86 / 0.79 & - & - & - & - & - & - \\
& PGD~\cite{pgd} & 0.76 / 0.58 / 0.16 & 0.84 / 0.83 / 0.72 & 0.03 / 0.03 / 0.07 & 0.00\% & 0.84 / 0.83 / 0.72 & 0.03 / 0.03 / 0.07 & 0.00\% \\
& C\&W~\cite{cw-l2} & 0.75 / 0.58 / 0.16 & 0.84 / 0.83 / 0.72 & 0.03 / 0.03 / 0.07 & 0.00\% & 0.84 / 0.83 / 0.72 & 0.03 / 0.03 / 0.07 & 0.00\% \\
& James' Attack~\cite{James_attack} & 0.90 / 0.81 / 0.26 & 0.84 / 0.83 / 0.72 & 0.03 / 0.03 / 0.07 & 0.00\% & 0.84 / 0.83 / 0.72 & 0.03 / 0.03 / 0.07 & 0.00\% \\
& AdvGPS~\cite{advgps} & 0.50 / 0.47 / 0.35 & 0.64 / 0.62 / 0.48 & 0.23 / 0.24 / 0.31 & 67.90\% & 0.84 / 0.83 / 0.72 & 0.03 / 0.03 / 0.07 & 0.00\% \\
\rowcolor{gray!12}
& \textbf{PB(Ours)} & 0.58 / 0.56 / 0.48 & \textbf{0.59 / 0.57 / 0.48} & \textbf{0.28 / 0.29 / 0.31} & \textbf{96.10\%} & \textbf{0.59 / 0.57 / 0.48} & \textbf{0.28 / 0.29 / 0.31} & \textbf{92.50\%} \\ \hline

\multirow{6}{*}{V2V4Real~\cite{V2v4real}}
& No Attack & 0.57 / 0.53 / 0.30 & - & - & - & - & - & - \\
& PGD~\cite{pgd} & 0.50 / 0.41 / 0.06 & 0.56 / 0.51 / 0.28 & 0.01 / 0.02 / 0.02 & 0.00\% & 0.57 / 0.51 / 0.28 & 0.00 / 0.02 / 0.02 & 0.00\% \\
& C\&W~\cite{cw-l2} & 0.36 / 0.31 / 0.07 & 0.56 / 0.51 / 0.28 & 0.01 / 0.02 / 0.02 & 0.00\% & 0.57 / 0.51 / 0.28 & 0.00 / 0.02 / 0.02 & 0.00\% \\
& James' Attack~\cite{James_attack} & 0.47 / 0.27 / 0.01 & 0.56 / 0.51 / 0.28 & 0.01 / 0.02 / 0.02 & 0.00\% & 0.57 / 0.51 / 0.28 & 0.00 / 0.02 / 0.02 & 0.00\% \\
& AdvGPS~\cite{advgps} & 0.19 / 0.15 / 0.08 & 0.49 / 0.46 / 0.26 & 0.08 / 0.07 / 0.04 & 34.20\% & 0.57 / 0.51 / 0.28 & 0.00 / 0.02 / 0.02 & 0.00\% \\
\rowcolor{gray!12}
& \textbf{PB(Ours)} & 0.35 / 0.32 / 0.16 & \textbf{0.35 / 0.32 / 0.16} & \textbf{0.22 / 0.21 / 0.14} & \textbf{98.50\%} & \textbf{0.41 / 0.38 / 0.21} & \textbf{0.16 / 0.15 / 0.09} & \textbf{85.70\%} \\ \hline

\end{tabular}
}
\label{tab:whitebox_attack}
\end{table*}

\begin{table*}[t]
\caption{Performance of PB, TrPB, and RT-TrPB across three datasets using PointPillars~\cite{pointpillars} + V2VNet~\cite{v2vnet} as the victim model (unseen for black-box attacks). Time Cost (ms) denotes adversarial perturbation generation time only.}
\centering
\setlength{\tabcolsep}{3.5pt}
\resizebox{\textwidth}{!}{
\begin{tabular}{cccccccccc}
\hline
\multirow{3}{*}{Dataset}
& \multirow{3}{*}{Attack}
& \multirow{2}{*}{No Defense}
& \multicolumn{3}{c}{\multirow{2}{*}{ROBOSAC Defense~\cite{robosac}}}
& \multicolumn{3}{c}{\multirow{2}{*}{FLD Defense~\cite{pb}}}
& \multirow{3}{*}{\begin{tabular}[c]{@{}c@{}}Time \\ Cost (ms)\end{tabular}} \\
& & & \multicolumn{3}{c}{} & \multicolumn{3}{c}{} & \\
\cmidrule(lr){3-3}\cmidrule(lr){4-6}\cmidrule(lr){7-9}
& & AP@(0.3 / 0.5 / 0.7) 
& AP@(0.3 / 0.5 / 0.7) $\downarrow$
& $\Delta$AP@(0.3 / 0.5 / 0.7) $\uparrow$
& DBR $\uparrow$
& AP@(0.3 / 0.5 / 0.7) $\downarrow$
& $\Delta$AP@(0.3 / 0.5 / 0.7) $\uparrow$
& DBR $\uparrow$
& \\ \hline

\multirow{4}{*}{OPV2V~\cite{opv2v}}
& No Attack & 0.91 / 0.91 / 0.85 & - & - & - & - & - & - & - \\
& PB & 0.54 / 0.53 / 0.42
& 0.54 / 0.53 / 0.42 & 0.37 / 0.38 / 0.43 & \textbf{100.00\%}
& 0.54 / 0.53 / 0.42 & 0.37 / 0.38 / 0.43 & \textbf{100.00\%}
& 5148.34 \\
& TrPB & 0.66 / 0.65 / 0.54
& 0.66 / 0.65 / 0.54 & 0.25 / 0.26 / 0.31 & \textbf{100.00\%}
& 0.66 / 0.65 / 0.54 & 0.25 / 0.26 / 0.31 & \textbf{99.80\%}
& 5689.59 \\
& RT-TrPB & 0.61 / 0.61 / 0.56
& 0.61 / 0.61 / 0.56 & 0.30 / 0.30 / 0.29 & \textbf{100.00\%}
& 0.61 / 0.61 / 0.56 & 0.30 / 0.30 / 0.29 & \textbf{97.60\%}
& \textbf{54.23} \\ \hline

\multirow{4}{*}{V2XSet~\cite{v2xset}}
& No Attack & 0.87 / 0.86 / 0.79 & - & - & - & - & - & - & - \\
& PB & 0.58 / 0.56 / 0.48
& 0.59 / 0.57 / 0.48 & 0.28 / 0.29 / 0.31 & \textbf{96.10\%}
& 0.59 / 0.57 / 0.48 & 0.28 / 0.29 / 0.31 & \textbf{92.50\%}
& 1348.06 \\
& TrPB & 0.74 / 0.71 / 0.53
& 0.74 / 0.72 / 0.53 & 0.13 / 0.14 / 0.26 & \textbf{98.60\%}
& 0.75 / 0.73 / 0.55 & 0.12 / 0.13 / 0.24 & \textbf{87.90\%}
& 2914.25 \\
& RT-TrPB & 0.77 / 0.76 / 0.65
& 0.78 / 0.77 / 0.65 & 0.09 / 0.09 / 0.14 & \textbf{96.10\%}
& 0.79 / 0.78 / 0.65 & 0.08 / 0.08 / 0.14 & \textbf{82.10\%}
& \textbf{55.54} \\ \hline

\multirow{4}{*}{V2V4Real~\cite{V2v4real}}
& No Attack & 0.57 / 0.53 / 0.30 & - & - & - & - & - & - & - \\
& PB & 0.35 / 0.32 / 0.16
& 0.35 / 0.32 / 0.16 & 0.22 / 0.21 / 0.14 & \textbf{98.50\%}
& 0.41 / 0.38 / 0.21 & 0.16 / 0.15 / 0.09 & \textbf{85.70\%}
& 1043.71 \\
& TrPB & 0.46 / 0.40 / 0.17
& 0.48 / 0.41 / 0.17 & 0.09 / 0.12 / 0.13 & \textbf{94.24\%}
& 0.48 / 0.42 / 0.18 & 0.09 / 0.11 / 0.12 & \textbf{91.89\%}
& 2382.81 \\
& RT-TrPB & 0.44 / 0.35 / 0.11
& 0.45 / 0.36 / 0.11 & 0.12 / 0.17 / 0.19 & \textbf{97.77\%}
& 0.45 / 0.35 / 0.11 & 0.12 / 0.18 / 0.19 & \textbf{100.00\%}
& \textbf{40.18} \\ \hline
\end{tabular}
}
\label{tab:blackbox_datasets}
\end{table*}

\subsection{Datasets}
We conduct experiments on three cooperative perception datasets: OPV2V~\cite{opv2v}, V2XSet~\cite{v2xset}, and V2V4Real~\cite{V2v4real}. OPV2V and V2XSet are large simulated datasets supporting vehicle-to-vehicle (V2V)~\cite{v2vnet} and vehicle-to-everything (V2X)~\cite{di-v2x} cooperation, respectively. Both contain complex driving scenarios with up to seven cooperating agents, while V2XSet also models realistic noise. In contrast, V2V4Real is a real-world V2V cooperative perception dataset involving two vehicles, including the ego agent, providing a practical benchmark for evaluating robustness under real-world sensing conditions. Across these datasets, we measure detection performance using average precision (AP) at IoU thresholds of 0.3, 0.5, and 0.7. Attack effectiveness and stealthiness are assessed by AP degradation and the defense-bypass rate (DBR), respectively, while defense effectiveness is measured by AP recovery under adversarial attacks.

\subsection{Implementation Details}

\textit{Cooperative Perception Models:}
We adopt two widely used 3D detection backbones, PointPillars~\cite{pointpillars} and SECOND~\cite{second}, combined with three feature fusion strategies, V2VNet~\cite{v2vnet}, AttFusion~\cite{opv2v}, and F-Cooper~\cite{F-cooper}, yielding six cooperative perception model configurations. All models are trained on the respective training splits, while perception evaluation and all attack and defense experiments are conducted on the validation splits. All training and evaluation experiments, including adversarial attacks and defenses, are performed on a single NVIDIA RTX 3090 GPU.

\textit{Adversarial Attack Settings:}
For white-box evaluation, we adopt PGD~\cite{pgd}, C\&W~\cite{cw-l2}, James' Attack~\cite{James_attack}, and AdvGPS~\cite{advgps} as baselines, with a perturbation budget of $\epsilon=0.5$. PGD, James' Attack, and AdvGPS use 15 optimization iterations, while C\&W uses 30. For PB, we use 30 iterations, set $\lambda_c=1.0$, and adopt region-specific perturbation budgets of $(0.1, 0.5, 0.1, 0.5, 0.3)$ for HCR, TAR, FAR, TIR, and FIR, respectively. When PointPillars + V2VNet serves as the victim model, we set $\mu_h=0.7$ and $\mu_l=0.3$. For black-box evaluation, TrPB inherits the PB settings and employs five heterogeneous surrogates in a leave-one-out setting. Its shared perturbation is optimized using Adam for 30 iterations with a learning rate of 0.025, and regularization weights $\lambda_{\mathrm{mom}}=0.05$ and $\lambda_{\mathrm{norm}}=0.001$.
For RT-TrPB, TrPB-generated region masks and perturbations from the training split are used to supervise the FRN and construct the RPD, respectively. The FRN is trained using AdamW for 8 epochs with a learning rate of 0.001 and a batch size of 8, adopting top-$K$ values of $(20,300,500,500)$ for HCR, TAR, FAR, and TIR. The RPD uses five shape bins and two orientation bins with sign-mean aggregation. During inference, RT-TrPB shares the perturbation budgets of TrPB without iterative optimization.

\textit{Defense Settings:}
We adopt ROBOSAC~\cite{robosac}, FLD~\cite{pb}, and GLST~\cite{GLST} as defense baselines, following their original configurations. For STTrust, FCV uses thresholds of $\tau_f=0.2$, $\tau_{\mathrm{sim}}=0.5$, and $\tau_{\mathrm{IoU}}=0.1$. MTC employs 8-connected components with a minimum area of 3 cells, a maximum centroid matching distance of 12 cells, and a motion history of 3 frames, with $\tau_{\mathrm{hard}}=0.30$ and $\tau_{\mathrm{soft}}=0.45$. For HIT, we set $\rho_{\mathrm{init}}=0$, $N_{\mathrm{WU}}=5$, and $\tau_{\mathrm{trust}}=0.8$, with trust penalties $\lambda_S=0.1$ and $\lambda_H=0.5$. The historical window length is $W=11$, with $B_{\mathrm{PD}}=2$ and $B_{\mathrm{BL}}=20$, while recovery requires $N_{\mathrm{rec}}=4$ consecutive \textit{GOOD} observations.

\begin{table*}[ht]
\caption{Performance of PB, TrPB, and RT-TrPB across five additional victim models on OPV2V~\cite{opv2v}. Together with PointPillars~\cite{pointpillars} + V2VNet~\cite{v2vnet} in Table~\ref{tab:blackbox_datasets}, six leave-one-out settings are evaluated, each using the other five models as surrogates. Time Cost (ms) denotes adversarial perturbation generation time only.}
\centering
\setlength{\tabcolsep}{3.5pt}
\resizebox{\textwidth}{!}{
\begin{tabular}{cccccccccc}
\hline
\multirow{3}{*}{Victim Model}                                                        & \multirow{3}{*}{Attack} & \multirow{2}{*}{No Defense} & \multicolumn{3}{c}{\multirow{2}{*}{ROBOSAC Defense~\cite{robosac}}} & \multicolumn{3}{c}{\multirow{2}{*}{FLD Defense~\cite{pb}}}                        & \multirow{3}{*}{\begin{tabular}[c]{@{}c@{}}Time \\ Cost (ms)\end{tabular}} \\
 &                         &                             & \multicolumn{3}{c}{}                                 & \multicolumn{3}{c}{}                                                    &                                                                            \\
 \cmidrule(lr){3-3}\cmidrule(lr){4-6}\cmidrule(lr){7-9}
 &                         & AP@(0.3 / 0.5 / 0.7)           & AP@(0.3 / 0.5 / 0.7) $\downarrow$  & $\Delta$AP@(0.3 / 0.5 / 0.7) $\uparrow$  & DBR $\uparrow$       & AP@(0.3 / 0.5 / 0.7) $\downarrow$ & $\Delta$AP@(0.3 / 0.5 / 0.7) $\uparrow$ & \multicolumn{1}{c}{DBR $\uparrow$}        &                                                                            \\ \hline

\multirow{4}{*}{\begin{tabular}[c]{@{}c@{}}PointPillars~\cite{pointpillars} \\ + AttFusion~\cite{opv2v}\end{tabular}}
 & No Attack               & 0.90 / 0.90 / 0.79              & -                  & -                    & -         & -                 & -                   & \multicolumn{1}{c}{-}          & -                                                                          \\
 & PB                      & 0.65 / 0.62 / 0.52              & 0.48 / 0.46 / 0.34     & 0.42 / 0.44 / 0.45       & 99.70\%   & 0.61 / 0.60 / 0.51    & 0.29 / 0.30 / 0.28      & \multicolumn{1}{c}{100.00\%}  & 6746.04                                                                    \\
 & TrPB                    & 0.72 / 0.67 / 0.48              & 0.59 / 0.54 / 0.39     & 0.31 / 0.36 / 0.40       & 99.90\%   & 0.72 / 0.67 / 0.48    & 0.18 / 0.23 / 0.31      & \multicolumn{1}{c}{100.00\%}  & 4810.71                                                                    \\
 & RT-TrPB                 & 0.73 / 0.72 / 0.57              & 0.67 / 0.66 / 0.46     & 0.23 / 0.24 / 0.33       & 100.00\%  & 0.73 / 0.72 / 0.57    & 0.17 / 0.18 / 0.22      & \multicolumn{1}{c}{100.00\%}  & \textbf{53.20}                                                             \\ \hline

\multirow{4}{*}{\begin{tabular}[c]{@{}c@{}}PointPillars~\cite{pointpillars} \\ + F-Cooper~\cite{F-cooper}\end{tabular}}
 & No Attack               & 0.98 / 0.98 / 0.92              & -                  & -                    & -         & -                 & -                   & \multicolumn{1}{c}{-}          & -                                                                          \\
 & PB                      & 0.50 / 0.49 / 0.40              & 0.50 / 0.49 / 0.40     & 0.48 / 0.49 / 0.52       & 99.90\%   & 0.50 / 0.49 / 0.40    & 0.48 / 0.49 / 0.52      & \multicolumn{1}{c}{99.60\%}   & 5342.71                                                                    \\
 & TrPB                    & 0.61 / 0.60 / 0.49              & 0.61 / 0.60 / 0.49     & 0.37 / 0.38 / 0.43       & 100.00\%  & 0.61 / 0.60 / 0.49    & 0.37 / 0.38 / 0.43      & \multicolumn{1}{c}{99.60\%}   & 5658.35                                                                    \\
 & RT-TrPB                 & 0.70 / 0.57 / 0.45              & 0.70 / 0.57 / 0.45     & 0.28 / 0.41 / 0.47       & 100.00\%  & 0.69 / 0.57 / 0.45    & 0.29 / 0.41 / 0.47      & \multicolumn{1}{c}{100.00\%}  & \textbf{40.29}                                                             \\ \hline

\multirow{4}{*}{\begin{tabular}[c]{@{}c@{}}SECOND~\cite{second} \\ + V2VNet~\cite{v2vnet}\end{tabular}}
 & No Attack               & 0.94 / 0.94 / 0.91              & -                  & -                    & -         & -                 & -                   & \multicolumn{1}{c}{-}          & -                                                                          \\
 & PB                      & 0.33 / 0.33 / 0.28              & 0.33 / 0.33 / 0.28     & 0.61 / 0.61 / 0.63       & 100.00\%  & 0.34 / 0.33 / 0.28    & 0.60 / 0.61 / 0.63      & \multicolumn{1}{c}{99.60\%}   & 6973.28                                                                    \\
 & TrPB                    & 0.74 / 0.71 / 0.43              & 0.74 / 0.71 / 0.43     & 0.20 / 0.23 / 0.48       & 100.00\%  & 0.74 / 0.71 / 0.43    & 0.20 / 0.23 / 0.48      & \multicolumn{1}{c}{100.00\%}  & 5655.09                                                                    \\
 & RT-TrPB                 & 0.84 / 0.81 / 0.65              & 0.84 / 0.81 / 0.65     & 0.10 / 0.13 / 0.26       & 100.00\%  & 0.84 / 0.81 / 0.65    & 0.10 / 0.13 / 0.26      & \multicolumn{1}{c}{100.00\%}  & \textbf{47.44}                                                             \\ \hline

\multirow{4}{*}{\begin{tabular}[c]{@{}c@{}}SECOND~\cite{second}\\  + AttFusion~\cite{opv2v}\end{tabular}}
 & No Attack               & 0.98 / 0.98 / 0.94              & -                  & -                    & -         & -                 & -                   & \multicolumn{1}{c}{-}          & -                                                                          \\
 & PB                      & 0.16 / 0.16 / 0.14              & 0.50 / 0.48 / 0.44     & 0.48 / 0.50 / 0.50       & 97.80\%   & 0.61 / 0.60 / 0.56    & 0.37 / 0.38 / 0.38      & \multicolumn{1}{c}{93.50\%}   & 8867.14                                                                    \\
 & TrPB                    & 0.79 / 0.73 / 0.59              & 0.69 / 0.65 / 0.54     & 0.29 / 0.33 / 0.40       & 100.00\%  & 0.79 / 0.73 / 0.59    & 0.19 / 0.25 / 0.35      & \multicolumn{1}{c}{100.00\%}  & 4791.23                                                                    \\
 & RT-TrPB                 & 0.91 / 0.83 / 0.53              & 0.84 / 0.74 / 0.51     & 0.14 / 0.24 / 0.43       & 99.90\%   & 0.91 / 0.83 / 0.53    & 0.07 / 0.15 / 0.41      & \multicolumn{1}{c}{100.00\%}  & \textbf{45.45}                                                             \\ \hline

\multirow{4}{*}{\begin{tabular}[c]{@{}c@{}}SECOND~\cite{second} \\ + F-Cooper~\cite{F-cooper}\end{tabular}}
 & No Attack               & 0.98 / 0.98 / 0.95              & -                  & -                    & -         & -                 & -                   & \multicolumn{1}{c}{-}          & -                                                                          \\
 & PB                      & 0.21 / 0.21 / 0.20              & 0.50 / 0.48 / 0.37     & 0.48 / 0.50 / 0.58       & 98.60\%   & 0.22 / 0.21 / 0.20    & 0.76 / 0.77 / 0.75      & \multicolumn{1}{c}{97.70\%}   & 7248.24                                                                    \\
 & TrPB                    & 0.60 / 0.59 / 0.57              & 0.60 / 0.59 / 0.57     & 0.38 / 0.39 / 0.38       & 100.00\%  & 0.60 / 0.59 / 0.57    & 0.38 / 0.39 / 0.38      & \multicolumn{1}{c}{100.00\%}  & 5647.88                                                                    \\
 & RT-TrPB                 & 0.60 / 0.58 / 0.46              & 0.60 / 0.58 / 0.46     & 0.38 / 0.40 / 0.49       & 100.00\%  & 0.60 / 0.58 / 0.46    & 0.38 / 0.40 / 0.49      & \multicolumn{1}{c}{100.00\%}  & \textbf{42.98}                                                             \\ \hline
\end{tabular}
}
\label{tab:blackbox_models}
\end{table*}

\begin{table*}[ht]
\caption{Defense performance against PB, TrPB, and RT-TrPB on three datasets using PointPillars~\cite{pointpillars} + V2VNet~\cite{v2vnet} as the victim model (unseen for black-box attacks). AP Recovery is computed as $(AP_{\mathrm{def}}-AP_{\mathrm{no\ def}})/(AP_{\mathrm{UB}}-AP_{\mathrm{no\ def}})\times100\%$, where Upper Bound (UB) represents the maximum AP after ideal malicious collaborator rejection. Time Cost (ms) denotes end-to-end inference time.}
\centering
\setlength{\tabcolsep}{2pt}
\resizebox{\textwidth}{!}{
\begin{tabular}{ccccccccc}
\hline
\multirow{3}{*}{Dataset}  & \multirow{3}{*}{Defense} & \multicolumn{2}{c}{\multirow{2}{*}{PB Attack}} & \multicolumn{2}{c}{\multirow{2}{*}{TrPB Attack}} & \multicolumn{2}{c}{\multirow{2}{*}{RT-TrPB Attack}} & \multirow{3}{*}{\begin{tabular}[c]{@{}c@{}}Time \\ Cost (ms)\end{tabular}} \\
 & & \multicolumn{2}{c}{} & \multicolumn{2}{c}{} & \multicolumn{2}{c}{} & \\
\cmidrule(lr){3-4}\cmidrule(lr){5-6}\cmidrule(lr){7-8}
 & & AP@(0.3 / 0.5 / 0.7) $\uparrow$ & AP Recovery (\%) $\uparrow$ & AP@(0.3 / 0.5 / 0.7) $\uparrow$ & AP Recovery (\%) $\uparrow$ & AP@(0.3 / 0.5 / 0.7) $\uparrow$ & AP Recovery (\%) $\uparrow$ & \\ \hline
\multirow{6}{*}{OPV2V~\cite{opv2v}} & No Defense & 0.54 / 0.53 / 0.42 & - & 0.66 / 0.65 / 0.54 & - & 0.61 / 0.61 / 0.56 & - & 32.23 \\
 & Upper Bound & 0.90 / 0.89 / 0.80 & - & 0.90 / 0.89 / 0.80 & - & 0.90 / 0.89 / 0.80 & - & - \\
 & ROBOSAC~\cite{robosac} & 0.54 / 0.53 / 0.42 & 0.00 / 0.00 / 0.00 & 0.66 / 0.65 / 0.54 & 0.00 / 0.00 / 0.00 & 0.61 / 0.61 / 0.56 & 0.00 / 0.00 / 0.00 & 40.61 \\
 & FLD~\cite{pb} & 0.54 / 0.53 / 0.42 & 0.00 / 0.00 / 0.00 & 0.66 / 0.65 / 0.54 & 0.00 / 0.00 / 0.00 & 0.61 / 0.61 / 0.56 & 0.00 / 0.00 / 0.00 & 40.27 \\
 & GLST~\cite{GLST} & 0.60 / 0.59 / 0.49 & 16.67 / 16.67 / 18.42 & 0.71 / 0.70 / 0.60 & 20.83 / 20.83 / 23.08 & 0.66 / 0.66 / 0.60 & 17.24 / 17.86 / 16.67 & 61.36 \\
\rowcolor{gray!12}
 & STTrust & 0.89 / 0.88 / 0.77 & \textbf{97.22 / 97.22 / 92.11} & 0.89 / 0.88 / 0.78 & \textbf{95.83 / 95.83 / 92.31} & 0.89 / 0.88 / 0.77 & \textbf{96.55 / 96.43 / 87.50} & 52.64 \\ \hline
\multirow{6}{*}{V2XSet~\cite{v2xset}} & No Defense & 0.58 / 0.56 / 0.48 & - & 0.74 / 0.71 / 0.53 & - & 0.77 / 0.76 / 0.65 & - & 25.72 \\
 & Upper Bound & 0.85 / 0.83 / 0.72 & - & 0.85 / 0.83 / 0.72 & - & 0.85 / 0.83 / 0.72 & - & - \\
 & ROBOSAC~\cite{robosac} & 0.59 / 0.57 / 0.48 & 3.70 / 3.70 / 0.00 & 0.74 / 0.72 / 0.53 & 0.00 / 8.33 / 0.00 & 0.78 / 0.77 / 0.65 & 12.50 / 14.29 / 0.00 & 33.72 \\
 & FLD~\cite{pb} & 0.59 / 0.57 / 0.48 & 3.70 / 3.70 / 0.00 & 0.75 / 0.73 / 0.55 & 9.09 / 16.67 / 10.53 & 0.79 / 0.78 / 0.65 & 25.00 / 28.57 / 0.00 & 30.22 \\
 & GLST~\cite{GLST} & 0.60 / 0.58 / 0.51 & 7.41 / 7.41 / 12.50 & 0.78 / 0.75 / 0.63 & 36.36 / 33.33 / 52.63 & 0.78 / 0.77 / 0.65 & 12.50 / 14.29 / 0.00 & 34.38 \\
\rowcolor{gray!12}
 & STTrust & 0.81 / 0.77 / 0.64 & \textbf{85.19 / 77.78 / 66.67} & 0.80 / 0.78 / 0.68 & \textbf{54.55 / 58.33 / 78.95} & 0.82 / 0.79 / 0.69 & \textbf{62.50 / 42.86 / 57.14} & 37.02 \\ \hline
\multirow{6}{*}{V2V4Real~\cite{V2v4real}} & No Defense & 0.35 / 0.32 / 0.16 & - & 0.46 / 0.40 / 0.17 & - & 0.44 / 0.35 / 0.11 & - & 16.64 \\
 & Upper Bound & 0.57 / 0.55 / 0.33 & - & 0.57 / 0.55 / 0.33 & - & 0.57 / 0.55 / 0.33 & - & - \\
 & ROBOSAC~\cite{robosac} & 0.35 / 0.32 / 0.16 & 0.00 / 0.00 / 0.00 & 0.48 / 0.41 / 0.17 & 18.18 / 6.67 / 0.00 & 0.45 / 0.36 / 0.11 & 7.69 / 5.00 / 0.00 & 21.67 \\
 & FLD~\cite{pb} & 0.41 / 0.38 / 0.21 & 27.27 / 26.09 / 29.41 & 0.48 / 0.42 / 0.18 & 18.18 / 13.33 / 6.25 & 0.45 / 0.35 / 0.11 & 7.69 / 0.00 / 0.00 & 19.12 \\
 & GLST~\cite{GLST} & 0.52 / 0.50 / 0.27 & 77.27 / 78.26 / 64.71 & 0.57 / 0.55 / 0.31 & 100.00 / 100.00 / 87.50 & 0.56 / 0.53 / 0.21 & 92.31 / 90.00 / 45.45 & 23.04 \\
\rowcolor{gray!12}
 & STTrust & 0.57 / 0.55 / 0.33 & \textbf{100.00 / 100.00 / 100.00} & 0.57 / 0.55 / 0.33 & \textbf{100.00 / 100.00 / 100.00} & 0.57 / 0.55 / 0.33 & \textbf{100.00 / 100.00 / 100.00} & 19.77 \\ \hline
\end{tabular}
}
\label{tab:sttrust_datasets}
\end{table*}

\subsection{Evaluation of Stealthy Adversarial Attacks}
\label{sec:attack_results}

\textit{White-Box Attack Effectiveness and Stealthiness:}
As shown in Table~\ref{tab:whitebox_attack}, PB achieves substantially higher defense-bypass rates than the baseline attacks while maintaining considerable perception degradation. Specifically, PB achieves DBR values of 100.00\%, 92.50\%, and 85.70\% against FLD~\cite{pb} on OPV2V, V2XSet, and V2V4Real, respectively, whereas all baseline attacks exhibit a DBR of 0\%. Meanwhile, PB reduces AP@0.5 from 0.91 to 0.53 on OPV2V even under FLD defense. This advantage stems from PB's collaboration-aware attack strategy, which preserves reliable consensus in high-confidence regions while selectively manipulating uncertain and ego-imperceptible regions. These results highlight the effectiveness of PB in jointly achieving perception degradation and attack stealthiness. Additional results across heterogeneous victim architectures are provided in the supplementary material.

\begin{table*}[ht]
\caption{Defense performance against PB, TrPB, and RT-TrPB on OPV2V~\cite{opv2v} across five additional victim models. Together with PointPillars~\cite{pointpillars} + V2VNet~\cite{v2vnet} in Table~\ref{tab:sttrust_datasets}, six victim configurations are evaluated (unseen for black-box attacks). Time Cost (ms) denotes End-to-end inference time. Negative recovery means that the defended AP is below the no-defense attacked AP.}
\centering
\setlength{\tabcolsep}{2pt}
\resizebox{\textwidth}{!}{
\begin{tabular}{ccccccccc}
\hline
\multirow{3}{*}{Victim Model} & \multirow{3}{*}{Defense} & \multicolumn{2}{c}{\multirow{2}{*}{PB Attack}} & \multicolumn{2}{c}{\multirow{2}{*}{TrPB Attack}} & \multicolumn{2}{c}{\multirow{2}{*}{RT-TrPB Attack}} & \multirow{3}{*}{\begin{tabular}[c]{@{}c@{}}Time \\ Cost (ms)\end{tabular}} \\
 & & \multicolumn{2}{c}{} & \multicolumn{2}{c}{} & \multicolumn{2}{c}{} & \\
\cmidrule(lr){3-4}\cmidrule(lr){5-6}\cmidrule(lr){7-8}
 & & AP@(0.3 / 0.5 / 0.7) $\uparrow$ & AP Recovery (\%) $\uparrow$ & AP@(0.3 / 0.5 / 0.7) $\uparrow$ & AP Recovery (\%) $\uparrow$ & AP@(0.3 / 0.5 / 0.7) $\uparrow$ & AP Recovery (\%) $\uparrow$ & \\ \hline
\multirow{6}{*}{\begin{tabular}[c]{@{}c@{}}PointPillars~\cite{pointpillars} \\ + AttFusion~\cite{opv2v}\end{tabular}} & No Defense & 0.65 / 0.62 / 0.52 & - & 0.72 / 0.67 / 0.48 & - & 0.73 / 0.72 / 0.57 & - & 44.54 \\
 & Upper Bound & 0.88 / 0.87 / 0.74 & - & 0.88 / 0.87 / 0.74 & - & 0.88 / 0.87 / 0.74 & - & - \\
 & ROBOSAC~\cite{robosac} & 0.48 / 0.46 / 0.34 & -73.91 / -64.00 / -81.82 & 0.59 / 0.54 / 0.39 & -81.25 / -65.00 / -34.62 & 0.67 / 0.66 / 0.46 & -40.00 / -40.00 / -64.71 & 53.14 \\
 & FLD~\cite{pb} & 0.61 / 0.60 / 0.51 & -17.39 / -8.00 / -4.55 & 0.72 / 0.67 / 0.48 & 0.00 / 0.00 / 0.00 & 0.73 / 0.72 / 0.57 & 0.00 / 0.00 / 0.00 & 52.66 \\
 & GLST~\cite{GLST} & 0.63 / 0.62 / 0.55 & -8.70 / 0.00 / 13.64 & 0.74 / 0.69 / 0.50 & 12.50 / 10.00 / 7.69 & 0.74 / 0.74 / 0.60 & 6.67 / 13.33 / 17.65 & 75.77 \\
\rowcolor{gray!12}
 & STTrust & 0.86 / 0.85 / 0.73 & \textbf{91.30 / 92.00 / 95.45} & 0.86 / 0.85 / 0.73 & \textbf{87.50 / 90.00 / 96.15} & 0.85 / 0.85 / 0.72 & \textbf{80.00 / 86.67 / 88.24} & 53.39 \\ \hline
\multirow{6}{*}{\begin{tabular}[c]{@{}c@{}}PointPillars~\cite{pointpillars} \\ + F-Cooper~\cite{F-cooper}\end{tabular}} & No Defense & 0.50 / 0.49 / 0.40 & - & 0.61 / 0.60 / 0.49 & - & 0.70 / 0.57 / 0.45 & - & 33.07 \\
 & Upper Bound & 0.96 / 0.94 / 0.86 & - & 0.96 / 0.94 / 0.86 & - & 0.96 / 0.94 / 0.86 & - & - \\
 & ROBOSAC~\cite{robosac} & 0.50 / 0.49 / 0.40 & 0.00 / 0.00 / 0.00 & 0.61 / 0.60 / 0.49 & 0.00 / 0.00 / 0.00 & 0.70 / 0.57 / 0.45 & 0.00 / 0.00 / 0.00 & 40.82 \\
 & FLD~\cite{pb} & 0.50 / 0.49 / 0.40 & 0.00 / 0.00 / 0.00 & 0.61 / 0.60 / 0.49 & 0.00 / 0.00 / 0.00 & 0.69 / 0.57 / 0.45 & -3.85 / 0.00 / 0.00 & 41.10 \\
 & GLST~\cite{GLST} & 0.50 / 0.50 / 0.31 & 0.00 / 2.22 / -19.57 & 0.61 / 0.60 / 0.44 & 0.00 / 0.00 / -13.51 & 0.70 / 0.58 / 0.41 & 0.00 / 2.70 / -9.76 & 63.05 \\
\rowcolor{gray!12}
 & STTrust & 0.92 / 0.91 / 0.82 & \textbf{91.30 / 93.33 / 91.30} & 0.92 / 0.91 / 0.82 & \textbf{88.57 / 91.18 / 89.19} & 0.86 / 0.82 / 0.72 & \textbf{61.54 / 67.57 / 65.85} & 50.18 \\ \hline
\multirow{6}{*}{\begin{tabular}[c]{@{}c@{}}SECOND~\cite{second}\\ + V2VNet~\cite{v2vnet}\end{tabular}} & No Defense & 0.33 / 0.33 / 0.28 & - & 0.74 / 0.71 / 0.43 & - & 0.84 / 0.81 / 0.65 & - & 54.79 \\
 & Upper Bound & 0.91 / 0.91 / 0.88 & - & 0.91 / 0.91 / 0.88 & - & 0.91 / 0.91 / 0.88 & - & - \\
 & ROBOSAC~\cite{robosac} & 0.33 / 0.33 / 0.28 & 0.00 / 0.00 / 0.00 & 0.74 / 0.71 / 0.43 & 0.00 / 0.00 / 0.00 & 0.84 / 0.81 / 0.65 & 0.00 / 0.00 / 0.00 & 62.94 \\
 & FLD~\cite{pb} & 0.34 / 0.33 / 0.28 & 1.72 / 0.00 / 0.00 & 0.74 / 0.71 / 0.43 & 0.00 / 0.00 / 0.00 & 0.84 / 0.81 / 0.65 & 0.00 / 0.00 / 0.00 & 61.84 \\
 & GLST~\cite{GLST} & 0.38 / 0.37 / 0.31 & 8.62 / 6.90 / 5.00 & 0.77 / 0.77 / 0.49 & 17.65 / 30.00 / 13.33 & 0.85 / 0.82 / 0.67 & 14.29 / 10.00 / 8.70 & 91.23 \\
\rowcolor{gray!12}
 & STTrust & 0.90 / 0.89 / 0.85 & \textbf{98.28 / 96.55 / 95.00} & 0.90 / 0.89 / 0.85 & \textbf{94.12 / 90.00 / 93.33} & 0.90 / 0.89 / 0.85 & \textbf{85.71 / 80.00 / 86.96} & 72.18 \\ \hline
\multirow{6}{*}{\begin{tabular}[c]{@{}c@{}}SECOND~\cite{second}\\ + AttFusion~\cite{opv2v}\end{tabular}} & No Defense & 0.16 / 0.16 / 0.14 & - & 0.79 / 0.73 / 0.59 & - & 0.91 / 0.83 / 0.53 & - & 68.67 \\
 & Upper Bound & 0.94 / 0.93 / 0.88 & - & 0.94 / 0.93 / 0.88 & - & 0.94 / 0.93 / 0.88 & - & - \\
 & ROBOSAC~\cite{robosac} & 0.50 / 0.48 / 0.44 & 43.59 / 41.56 / 40.54 & 0.69 / 0.65 / 0.54 & -66.67 / -40.00 / -17.24 & 0.84 / 0.74 / 0.51 & -233.33 / -90.00 / -5.71 & 164.35 \\
 & FLD~\cite{pb} & 0.61 / 0.60 / 0.56 & 57.69 / 57.14 / 56.76 & 0.79 / 0.73 / 0.59 & 0.00 / 0.00 / 0.00 & 0.91 / 0.83 / 0.53 & 0.00 / 0.00 / 0.00 & 77.14 \\
 & GLST~\cite{GLST} & 0.17 / 0.17 / 0.15 & 1.28 / 1.30 / 1.35 & 0.80 / 0.74 / 0.60 & 6.67 / 5.00 / 3.45 & 0.92 / 0.84 / 0.55 & 33.33 / 10.00 / 5.71 & 105.73 \\
\rowcolor{gray!12}
 & STTrust & 0.92 / 0.91 / 0.85 & \textbf{97.44 / 97.40 / 95.95} & 0.92 / 0.91 / 0.85 & \textbf{86.67 / 90.00 / 89.66} & 0.92 / 0.90 / 0.77 & \textbf{33.33 / 70.00 / 68.57} & 88.54 \\ \hline
\multirow{6}{*}{\begin{tabular}[c]{@{}c@{}}SECOND~\cite{second} \\ + F-Cooper~\cite{F-cooper}\end{tabular}} & No Defense & 0.21 / 0.21 / 0.20 & - & 0.60 / 0.59 / 0.57 & - & 0.60 / 0.58 / 0.46 & - & 55.52 \\
 & Upper Bound & 0.95 / 0.94 / 0.90 & - & 0.95 / 0.94 / 0.90 & - & 0.95 / 0.94 / 0.90 & - & - \\
 & ROBOSAC~\cite{robosac} & 0.50 / 0.48 / 0.37 & 39.19 / 36.99 / 24.29 & 0.60 / 0.59 / 0.57 & 0.00 / 0.00 / 0.00 & 0.60 / 0.58 / 0.46 & 0.00 / 0.00 / 0.00 & 105.81 \\
 & FLD~\cite{pb} & 0.22 / 0.21 / 0.20 & 1.35 / 0.00 / 0.00 & 0.60 / 0.59 / 0.57 & 0.00 / 0.00 / 0.00 & 0.60 / 0.58 / 0.46 & 0.00 / 0.00 / 0.00 & 65.50 \\
 & GLST~\cite{GLST} & 0.22 / 0.22 / 0.20 & 1.35 / 1.37 / 0.00 & 0.60 / 0.59 / 0.56 & 0.00 / 0.00 / -3.03 & 0.60 / 0.58 / 0.45 & 0.00 / 0.00 / -2.27 & 92.73 \\
\rowcolor{gray!12}
 & STTrust & 0.90 / 0.89 / 0.83 & \textbf{93.24 / 93.15 / 90.00} & 0.90 / 0.89 / 0.84 & \textbf{85.71 / 85.71 / 81.82} & 0.90 / 0.89 / 0.83 & \textbf{85.71 / 86.11 / 84.09} & 83.24 \\ \hline
\end{tabular}
}
\label{tab:sttrust_models}
\end{table*}

\textit{Zero-Query Black-Box Attacks Across Datasets:}
As shown in Table~\ref{tab:blackbox_datasets}, we evaluate TrPB and RT-TrPB under a zero-query black-box setting, using five heterogeneous surrogate models and holding out PointPillars~\cite{pointpillars} with V2VNet~\cite{v2vnet} as the unseen victim. TrPB maintains attack effectiveness and stealthiness across all three datasets, achieving FLD~\cite{pb} DBR values of 99.80\%, 87.90\%, and 91.89\% on OPV2V~\cite{opv2v}, V2XSet~\cite{v2xset}, and V2V4Real~\cite{V2v4real}, respectively, while reducing AP@0.5 to 0.65, 0.73, and 0.42. Although its attack strength is generally lower than white-box PB, these results indicate that TrPB can transfer stealthy attack behaviors to an unseen victim through heterogeneous surrogate consensus and shared optimization. Furthermore, RT-TrPB substantially reduces perturbation generation time from 5689.59 to 54.23 ms on OPV2V, achieving a 104.9$\times$ speedup, with similarly significant improvements on the other datasets. This gain stems from replacing iterative surrogate optimization with lightweight region prediction and dictionary-based perturbation composition, while retaining substantial attack effectiveness despite dataset-dependent performance variations. 

\textit{Transferability Across Heterogeneous Victim Models:}
As shown in Table~\ref{tab:blackbox_models}, we evaluate five additional unseen victim models on OPV2V~\cite{opv2v}, complementing the PointPillars~\cite{pointpillars} + V2VNet~\cite{v2vnet} results in Table~\ref{tab:blackbox_datasets} to form six leave-one-out black-box settings across different perception backbones and fusion strategies. TrPB reduces AP@0.3 by 0.18--0.38 across these five unseen victim models while maintaining a FLD DBR of 99.60\%--100.00\%, demonstrating effective attack transferability without victim-specific knowledge. Similarly, RT-TrPB achieves a FLD DBR of 100.00\% across all five models, with perturbation generation times ranging from 40.29 to 53.20 ms. Despite variations in AP degradation across victim configurations, both methods maintain high defense-bypass rates while degrading perception performance. These results indicate that heterogeneous surrogate consensus and shared optimization enable TrPB to capture transferable attack behaviors, while RT-TrPB retains such capabilities through offline knowledge amortization. Overall, both methods demonstrate zero-query black-box transferability across heterogeneous victim models.

\begin{figure}[t]
  \centering
   \includegraphics[width=\linewidth]{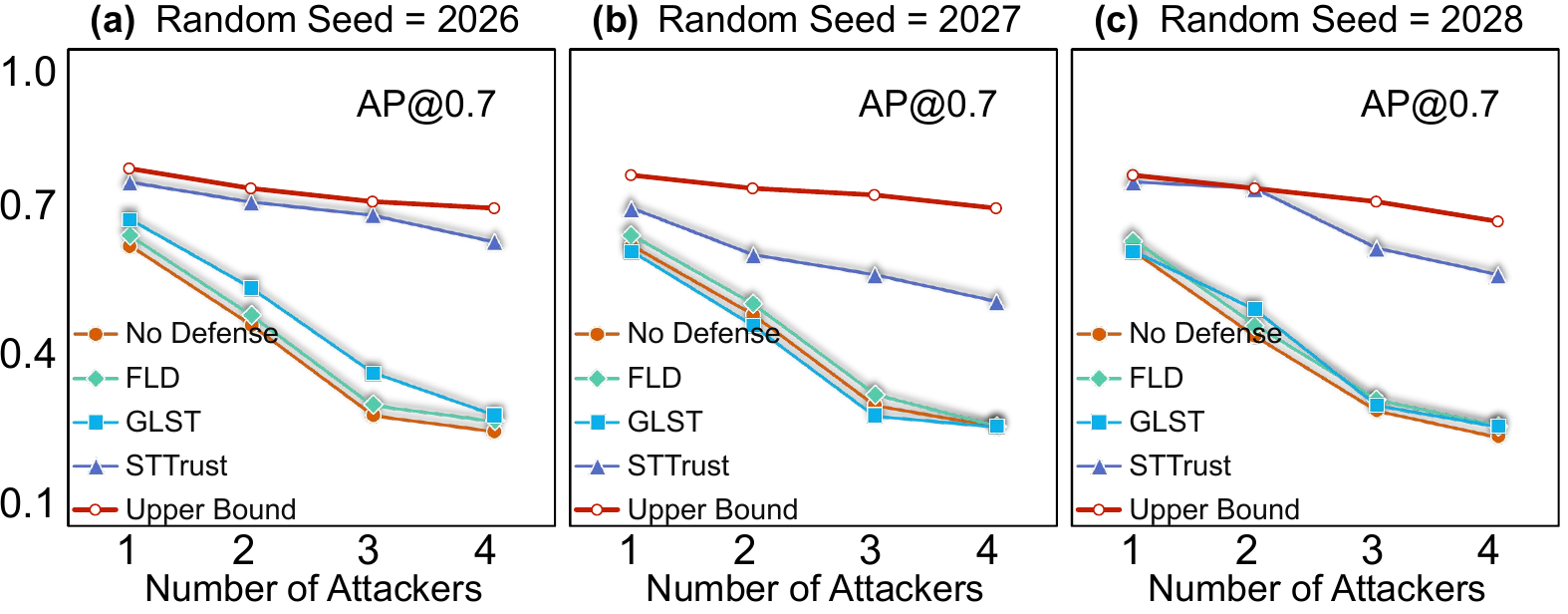}
   \caption{Performance of Different Defenses against PB with Increasing Attackers on OPV2V Using PointPillars with V2VNet as the Victim Model.}
   \label{fig:multi_attackers}
\end{figure}

\subsection{Evaluation of Spatio-Temporal Trust Defense}
\label{sec:defense_results}

\textit{Defense Robustness Across Datasets:}
As shown in Table~\ref{tab:sttrust_datasets}, STTrust achieves substantially higher AP Recovery than existing defenses against PB, TrPB, and RT-TrPB across all three datasets. On OPV2V~\cite{opv2v}, STTrust achieves AP Recovery rates of 97.22\%, 95.83\%, and 96.43\% at IoU 0.5, respectively, whereas ROBOSAC~\cite{robosac} and FLD~\cite{pb} achieve 0\% against all three attacks. On V2V4Real~\cite{V2v4real}, STTrust reaches the upper-bound performance under all three attacks, while maintaining the highest AP Recovery on V2XSet~\cite{v2xset} despite some performance variations. These improvements highlight the limitations of conventional consistency-based defenses, such as ROBOSAC and FLD, which primarily rely on instantaneous collaborative agreement and struggle with attacks that deliberately preserve benign-like consistency. Although GLST~\cite{GLST} incorporates trust estimation, its AP Recovery remains lower across the evaluated settings. In contrast, STTrust integrates feature consistency, motion-aware temporal coherence, and historical identity trust to assess collaborators beyond individual frames, providing more reliable defense against stealthy attacks with moderate inference overhead.

\begin{table}[t]
\caption{Ablation of PB region selection and attack objectives on OPV2V using PointPillars with V2VNet as the victim model.}
\centering
\setlength{\tabcolsep}{2pt}
\resizebox{\columnwidth}{!}{
\begin{tabular}{cccccc}
\hline
\multirow{2}{*}{\begin{tabular}[c]{@{}c@{}}Region \\ Selection\end{tabular}} & \multirow{2}{*}{\begin{tabular}[c]{@{}c@{}}Attack Objective \\ Optimization\end{tabular}} & \multicolumn{2}{c}{ROBOSAC Defense} & \multicolumn{2}{c}{FLD Defense} \\
\cmidrule(lr){3-4}\cmidrule(lr){5-6}
& & $\Delta$AP@(0.3/0.5/0.7) & DBR & $\Delta$AP@(0.3/0.5/0.7) & DBR \\
\hline
Random & FB Attack & 0.26 / 0.26 / 0.26 & 78.00\% & 0.07 / 0.08 / 0.09 & 36.00\% \\
LCS & FB Attack & 0.11 / 0.13 / 0.17 & 90.00\% & 0.02 / 0.03 / 0.06 & 40.00\% \\
ARS & FB Attack & 0.08 / 0.10 / 0.44 & 100.00\% & 0.05 / 0.07 / 0.32 & 69.00\% \\
ARS & MAPG w/o CPR & 0.46 / 0.48 / 0.49 & 100.00\% & 0.45 / 0.46 / 0.47 & 96.00\% \\
ARS & MAPG & 0.37 / 0.38 / 0.43 & 100.00\% & 0.37 / 0.38 / 0.43 & 100.00\% \\
\hline
\end{tabular}
}
\label{tab:pb_module_ablation}
\end{table}

\begin{table}[t]
\caption{Ablation of PB semantic regions on OPV2V~\cite{opv2v} using PointPillars~\cite{pointpillars} with V2VNet~\cite{v2vnet} as the victim model.}
\centering
\setlength{\tabcolsep}{6pt}
\resizebox{\columnwidth}{!}{
\begin{tabular}{ccccc}
\hline
\multirow{2}{*}{\begin{tabular}[c]{@{}c@{}}Excluded \\ Region\end{tabular}} & \multicolumn{2}{c}{ROBOSAC Defense~\cite{robosac}} & \multicolumn{2}{c}{FLD Defense~\cite{pb}} \\
\cmidrule(lr){2-3}\cmidrule(lr){4-5}
& $\Delta$AP@(0.3/0.5/0.7)       & DBR         & $\Delta$AP@(0.3/0.5/0.7)    & DBR       \\ \hline
none                            & 0.37 / 0.38 / 0.43     & 100.00\%    & 0.37 / 0.38 / 0.43  & 100.00\%  \\
HCR                             & 0.32 / 0.33 / 0.32     & 100.00\%    & 0.32 / 0.33 / 0.32  & 100.00\%  \\
TAR                             & 0.23 / 0.25 / 0.31     & 100.00\%    & 0.23 / 0.25 / 0.31  & 100.00\%  \\
FAR                             & 0.34 / 0.34 / 0.39     & 99.00\%     & 0.34 / 0.34 / 0.39  & 100.00\%  \\
TIR                             & 0.24 / 0.25 / 0.32     & 100.00\%    & 0.24 / 0.25 / 0.32  & 100.00\%  \\
FIR                             & 0.31 / 0.32 / 0.38     & 100.00\%    & 0.31 / 0.32 / 0.38  & 100.00\%  \\ \hline
\end{tabular}
}
\label{tab:pb_region_ablation}
\end{table}

\begin{table}[t]
\caption{Ablation of TrPB aggregation strategies on OPV2V~\cite{opv2v} using PointPillars~\cite{pointpillars} with V2VNet~\cite{v2vnet} as the victim model.}
\centering
\setlength{\tabcolsep}{2pt}
\resizebox{\columnwidth}{!}{
\begin{tabular}{ccccccc}
\hline
\multirow{2}{*}{\begin{tabular}[c]{@{}c@{}}Aggregation \\ Strategy\end{tabular}} & \multirow{2}{*}{\begin{tabular}[c]{@{}c@{}}Threshold \\ Calibration\end{tabular}} & \multicolumn{2}{c}{ROBOSAC Defense~\cite{robosac}} & \multicolumn{2}{c}{FLD Defense~\cite{pb}} & \multirow{2}{*}{\begin{tabular}[c]{@{}c@{}}Attack Area \\ Ratio (\%)\end{tabular}} \\
\cmidrule(lr){3-4}\cmidrule(lr){5-6}
&                                                                                   & $\Delta$AP@0.7       & DBR           & $\Delta$AP@0.7     & DBR         &                                                                                    \\ \hline
intersection                                                                     & Model-Specific                                                                    & 0.06                 & 100.00\%      & 0.06            & 100.00\%    & 0.22                                                                               \\
majority                                                                         & Model-Specific                                                                    & 0.28                 & 100.00\%      & 0.27            & 98.20\%     & 1.18                                                                               \\
union                                                                            & Model-Specific                                                                    & 0.36                 & 100.00\%      & 0.29            & 84.70\%     & 2.95                                                                               \\
TrPB proposed                                                                    & Model-Specific                                                                    & 0.31                 & 100.00\%      & 0.31            & 99.79\%     & 1.84                                                                               \\
TrPB proposed                                                                    & Shared (PP+V2VNet)                                                                & 0.24                 & 100.00\%      & 0.24            & 99.90\%     & 1.32                                                                               \\ \hline
\end{tabular}
}
\label{tab:trpb_aggregation_ablation}
\end{table}

\textit{Defense Robustness Across Perception Models:}
As shown in Table~\ref{tab:sttrust_models}, we further evaluate STTrust on five heterogeneous perception models on OPV2V~\cite{opv2v}, complementing the PointPillars~\cite{pointpillars} + V2VNet~\cite{v2vnet} results in Table~\ref{tab:sttrust_datasets}. Across these configurations, STTrust generally achieves higher AP Recovery than ROBOSAC~\cite{robosac}, FLD~\cite{pb}, and GLST~\cite{GLST} under all three attacks. At IoU 0.5, its AP Recovery ranges from 92.00\% to 97.40\% against PB, 85.71\% to 91.18\% against TrPB, and 67.57\% to 86.67\% against RT-TrPB. For instance, under RT-TrPB on PointPillars + AttFusion~\cite{opv2v}, STTrust achieves 86.67\% AP Recovery, compared with 13.33\% for GLST. These results demonstrate the effectiveness of STTrust across heterogeneous perception architectures, despite variations in recovery performance among different victim models.

\textit{Defense under Increasing Numbers of Attackers:}
Fig.~\ref{fig:multi_attackers} shows that STTrust consistently outperforms the other defenses across all three random seeds and achieves performance closest to the Upper Bound. As the number of attackers increases, the gap to the Upper Bound widens, but STTrust still maintains a clear advantage over the competing defenses.

\begin{figure}[t]
  \centering
   \includegraphics[width=\linewidth]{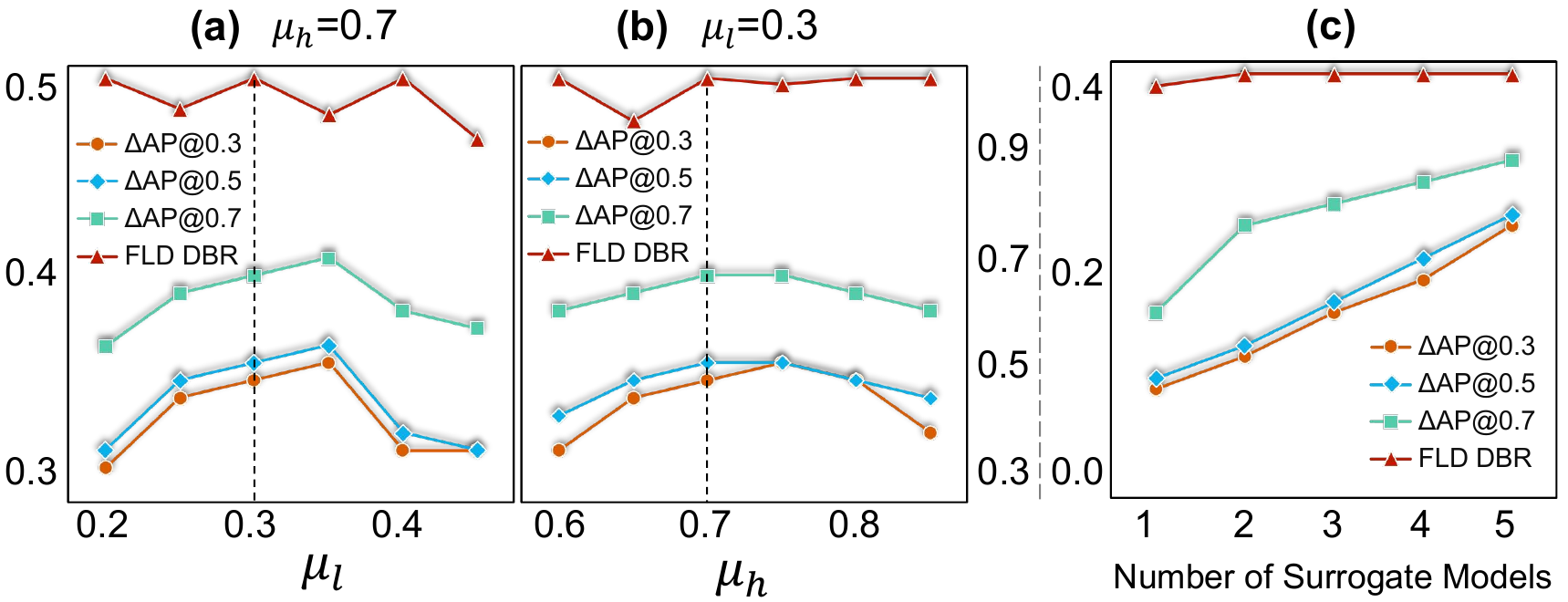}
    \caption{Hyperparameter analysis on OPV2V~\cite{opv2v} using PointPillars~\cite{pointpillars} with V2VNet~\cite{v2vnet} as the victim model. (a, b) Effects of PB confidence thresholds, with $\mu_h=0.7$ fixed in (a) and $\mu_l=0.3$ fixed in (b). (c) Effect of TrPB surrogate model pool size on attack performance.}
   \label{fig:pb_ablation_para}
\end{figure}

\begin{table}[t]
\caption{Ablation of TrPB surrogate model pool composition and size on OPV2V~\cite{opv2v}, averaged over six victim models.}
\centering
\setlength{\tabcolsep}{5pt}
\resizebox{0.95\columnwidth}{!}{
\begin{tabular}{ccccc}
\hline
\multirow{2}{*}{Surrogate Pool} & \multicolumn{2}{c}{ROBOSAC Defense~\cite{robosac}} & \multicolumn{2}{c}{FLD Defense~\cite{pb}} \\
\cmidrule(lr){2-3}\cmidrule(lr){4-5}
& $\Delta$AP@(0.3/0.5/0.7)         & DBR         & $\Delta$AP@(0.3/0.5/0.7)      & DBR       \\ \hline
cross-only-3                    & 0.16 / 0.21 / 0.27     & 99.00\%     & 0.11 / 0.18 / 0.25  & 92.00\%   \\
mixed-3                         & 0.22 / 0.28 / 0.34     & 100.00\%    & 0.19 / 0.27 / 0.33  & 100.00\%  \\
mixed-5                         & 0.30 / 0.32 / 0.40     & 100.00\%    & 0.26 / 0.29 / 0.38  & 100.00\%  \\ \hline
\end{tabular}
}
\label{tab:trpb_surrogate_pool_ablation}
\end{table}

\begin{table}[t]
\caption{Ablation of RT-TrPB modules on OPV2V~\cite{opv2v} using PointPillars~\cite{pointpillars} with V2VNet~\cite{v2vnet} as the victim model.}
\centering
\setlength{\tabcolsep}{3pt}
\resizebox{\columnwidth}{!}{
\begin{tabular}{lccccc}
\hline
\multirow{2}{*}{\begin{tabular}[c]{@{}c@{}}Attack\\  Configuration\end{tabular}} & \multicolumn{2}{c}{ROBOSAC Defense~\cite{robosac}} & \multicolumn{2}{c}{FLD Defense~\cite{pb}}     & \multirow{2}{*}{\begin{tabular}[c]{@{}c@{}}Time Cost\\  (ms)\end{tabular}} \\ \cmidrule(lr){2-3}\cmidrule(lr){4-5}
& $\Delta$AP@(0.3/0.5/0.7) & DBR      & $\Delta$AP@(0.3/0.5/0.7) & DBR      &                                                                            \\ \hline
\begin{tabular}[c]{@{}c@{}}Teacher Mask + \\ 30 Steps Opt.\end{tabular}          & 0.25 / 0.26 / 0.31       & 100.00\% & 0.25 / 0.26 / 0.31       & 99.79\%  & 5689.59                                                                    \\
\begin{tabular}[c]{@{}c@{}}Teacher Mask + \\ Dictionary\end{tabular}             & 0.24 / 0.25 / 0.31       & 100.00\% & 0.24 / 0.25 / 0.31       & 100.00\% & \textbf{726.98}                                                            \\
\begin{tabular}[c]{@{}c@{}}FRN Mask + \\ 30 Steps Opt.\end{tabular}              & 0.29 / 0.30 / 0.37       & 100.00\% & 0.24 / 0.25 / 0.31       & 82.38\%  & 5025.83                                                                    \\
\begin{tabular}[c]{@{}c@{}}FRN Mask + \\ Dictionary\end{tabular}                 & 0.30 / 0.30 / 0.29       & 100.00\% & 0.29 / 0.29 / 0.29       & 97.58\%  & \textbf{54.23}                                                             \\ \hline
\end{tabular}
}
\label{tab:rt_trpb_module_ablation}
\end{table}

\begin{table}[t]
\caption{Ablation of RT-TrPB input configurations under FLD defense on OPV2V using PointPillars with V2VNet as the victim model.}
\centering
\setlength{\tabcolsep}{2pt}
\resizebox{\columnwidth}{!}{
\begin{tabular}{lcccc}
\hline
& \multicolumn{3}{c}{FLD Defense~\cite{pb}} & \\
\cmidrule(lr){2-4}
\multirow{-2}{*}{Attack setup} & AP@(0.3 / 0.5 / 0.7) & $\Delta$AP@(0.3 / 0.5 / 0.7) & DBR & \multirow{-2}{*}{\begin{tabular}[c]{@{}c@{}}Time \\ Cost (ms)\end{tabular}} \\
\hline
RT-TrPB & 0.61 / 0.61 / 0.56 & 0.30 / 0.30 / 0.29 & 97.60\% & 54.23 \\
RT-TrPB (Attacker-only + pose) & 0.82 / 0.82 / 0.73 & 0.09 / 0.09 / 0.12 & 37.30\% & 53.34 \\
RT-TrPB (All attackers + pose) & 0.65 / 0.65 / 0.60 & 0.26 / 0.26 / 0.25 & 79.58\% & 54.08 \\
\hline
\end{tabular}
}
\label{tab:rt_trpb_info_ablation}
\end{table}

\subsection{Ablation Studies}
\label{sec:Ablation_Studies}

\textit{PB Region Selection and Attack Objective Ablation:}
We conduct ablation experiments on PB region selection and attack objectives, as shown in Table~\ref{tab:pb_module_ablation}. Compared with random area selection and low-confidence area selection (LCS), Attack Region Selection (ARS) increases $\Delta$AP@0.7 to 0.32 under FLD defense~\cite{pb}, demonstrating the importance of selecting appropriate attack regions for improving both attack effectiveness and stealthiness. Moreover, replacing the foreground-background swapping attack (FB Attack)~\cite{James_attack} with Multi-target Adversarial Perturbation Generation (MAPG), which assigns region-specific objectives to preserve reliable high-confidence responses, suppress likely true detections, and reinforce or fabricate false ones, further improves attack performance, increasing $\Delta$AP@0.7 to 0.47 and FLD DBR to 96.00\%. Furthermore, incorporating consensus-preserving regularization (CPR) raises DBR to 100.00\% while slightly reducing $\Delta$AP@0.7 from 0.47 to 0.43, indicating a better balance between perception degradation and stealthiness.

\textit{PB Semantic Region Ablation:}
We evaluate the contribution of each semantic region in PB by removing one region at a time, as shown in Table~\ref{tab:pb_region_ablation}. Removing the High-Confidence Region (HCR), True Ambiguous Region (TAR), or True Imperceptible Region (TIR) reduces $\Delta$AP@0.7 under FLD~\cite{pb} from 0.43 to 0.32, 0.31, and 0.32, respectively, while removing the False Ambiguous Region (FAR) or False Imperceptible Region (FIR) reduces it to 0.39 and 0.38. These results suggest that suppressing true perceptions contributes slightly more to perception degradation than reinforcing or fabricating false perceptions. Overall, these results demonstrate the complementary roles of preserving reliable consensus and manipulating both true and false perceptions in achieving effective and stealthy attacks.

\textit{TrPB Aggregation Strategy Ablation:}
We evaluate the Semantic Region Consensus (SRC) aggregation strategies in TrPB, as shown in Table~\ref{tab:trpb_aggregation_ablation}. Applying a single aggregation rule uniformly across regions fails to balance attack effectiveness and stealthiness. Intersection overly restricts the attack regions, resulting in a low $\Delta$AP@0.7 of 0.06, whereas union produces less precise regions and reduces the FLD~\cite{pb} DBR to 84.70\%. Majority voting performs better but remains inferior to our proposed strategy. Specifically, TrPB uses intersection for HCR to preserve cross-model consensus, majority voting for TAR to identify stable uncertain targets, and union for FAR and TIR to ensure sufficient attack coverage. This region-specific design achieves a $\Delta$AP@0.7 of 0.31 and an FLD DBR of 99.79\%. The supplementary material further analyzes the aggregation strategy for each region. We also compare model-specific and shared confidence thresholds: model-specific calibration increases $\Delta$AP@0.7 from 0.24 to 0.31 under FLD, while maintaining a similarly high DBR.

\textit{TrPB Surrogate Model Pool Ablation:}
We evaluate the impact of surrogate model pool composition and size on TrPB, with results averaged over six victim models, as shown in Table~\ref{tab:trpb_surrogate_pool_ablation}. Compared with cross-only-3, mixed-3 achieves higher $\Delta$AP@0.7 and DBR under both ROBOSAC~\cite{robosac} and FLD~\cite{pb} defenses, indicating that incorporating surrogate models with different backbone relationships to the victim can improve attack effectiveness and stealthiness. Increasing the surrogate pool from three to five models further raises $\Delta$AP@0.7 to 0.40 and 0.38 under ROBOSAC and FLD, respectively, while maintaining 100\% DBR. These results highlight the importance of surrogate pool composition and suggest that including diverse backbone architectures is beneficial for transferable attacks.

\textit{RT-TrPB Module Ablation:}
We conduct ablation experiments on the Fast Region Network (FRN) and Region Perturbation Dictionary (RPD) in RT-TrPB, as shown in Table~\ref{tab:rt_trpb_module_ablation}. Replacing Shared Constrained Optimization with RPD reduces the perturbation generation time from 5689.59 ms to 726.98 ms, while maintaining comparable attack effectiveness and DBR. Furthermore, replacing TrPB-derived attack masks with FRN-predicted masks and combining them with RPD reduces the generation time to only 54.23 ms. Compared with FRN masks combined with iterative optimization, this configuration also improves the FLD~\cite{pb} DBR from 82.38\% to 97.58\%. These results demonstrate that RPD substantially accelerates perturbation generation, while FRN enables further speedup and, when combined with RPD, improves attack reliability against the defense.

\textit{RT-TrPB Attack Configuration Ablation:}
We evaluate the impact of input availability on RT-TrPB, as shown in Table~\ref{tab:rt_trpb_info_ablation}. Restricting the input to an individual attacker's observations and relative poses substantially reduces $\Delta$AP@0.7 from 0.29 to 0.12 and the FLD~\cite{pb} DBR from 97.60\% to 37.30\%, highlighting the importance of collaborative information for effective attacks. Incorporating observations from all attackers partially alleviates this limitation, increasing $\Delta$AP@0.7 to 0.25 and the DBR to 79.58\%. These results demonstrate that full collaborative information plays an important role in RT-TrPB and that, when unavailable, observations from multiple attackers can partially compensate for its absence, helping maintain a degree of attack effectiveness and stealthiness.

\textit{STTrust Component Ablation:}
We conduct ablation experiments on the individual modules of STTrust, as shown in Table~\ref{tab:sttrust_module_ablation}. Feature Consistency Verification (FCV) alone achieves nearly 100\% benign collaborator acceptance (BA) but almost zero attack rejection (AR), indicating that single-frame consistency is insufficient to identify stealthy attacks. Adding Motion-aware Temporal Consistency (MTC) with a single-threshold binary decision (FCV + MTC (B)) substantially improves AR to 80.9\%--91.7\%, but reduces BA to around 71\%. Using MTC's dual-threshold scheme to retain graded temporal evidence improves BA to 94.0\%--95.9\%, but without Historical Identity Trust (HIT) to accumulate this evidence across frames, attack rejection remains limited, particularly against TrPB and RT-TrPB. Incorporating HIT further raises AR to 95.1\%--100.0\%, while BA remains around 69.7\%. These results demonstrate that historical identity-level evidence complements spatial and temporal consistency, substantially strengthening attack detection at the cost of reduced benign collaborator acceptance.

\begin{figure*}[ht]
  \centering
  \includegraphics[width=\linewidth]{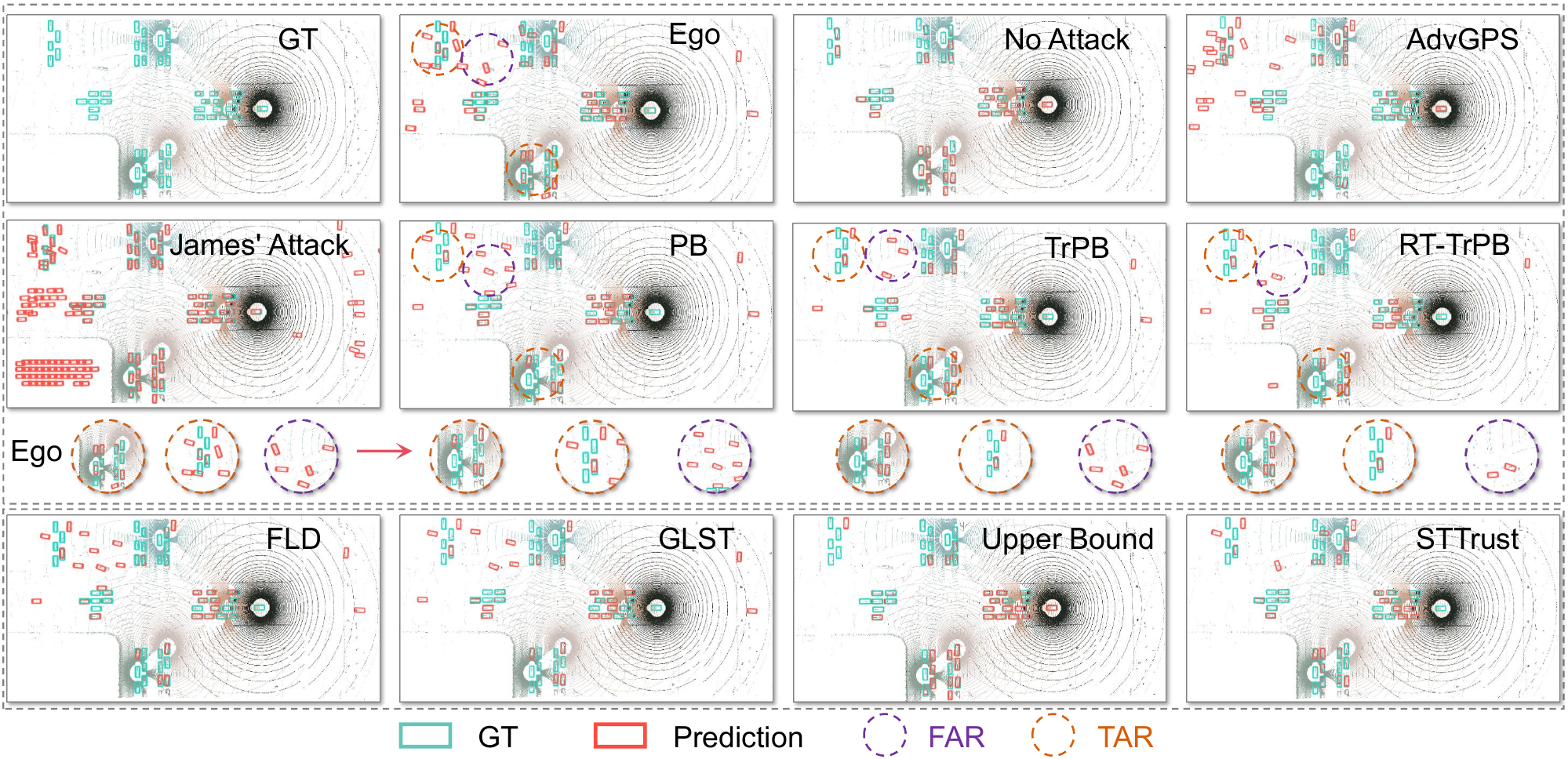}
    \caption{Qualitative comparison of attacks and defenses on OPV2V~\cite{opv2v} using PointPillars~\cite{pointpillars} with V2VNet~\cite{v2vnet} as the victim model. The upper panels compare attack outcomes and highlight the FAR and TAR regions, while the lower panels compare defense results against PB.}
  \label{fig:qualitative_main}
\end{figure*}

\begin{figure}[t]
  \centering
   \includegraphics[width=0.75\linewidth]{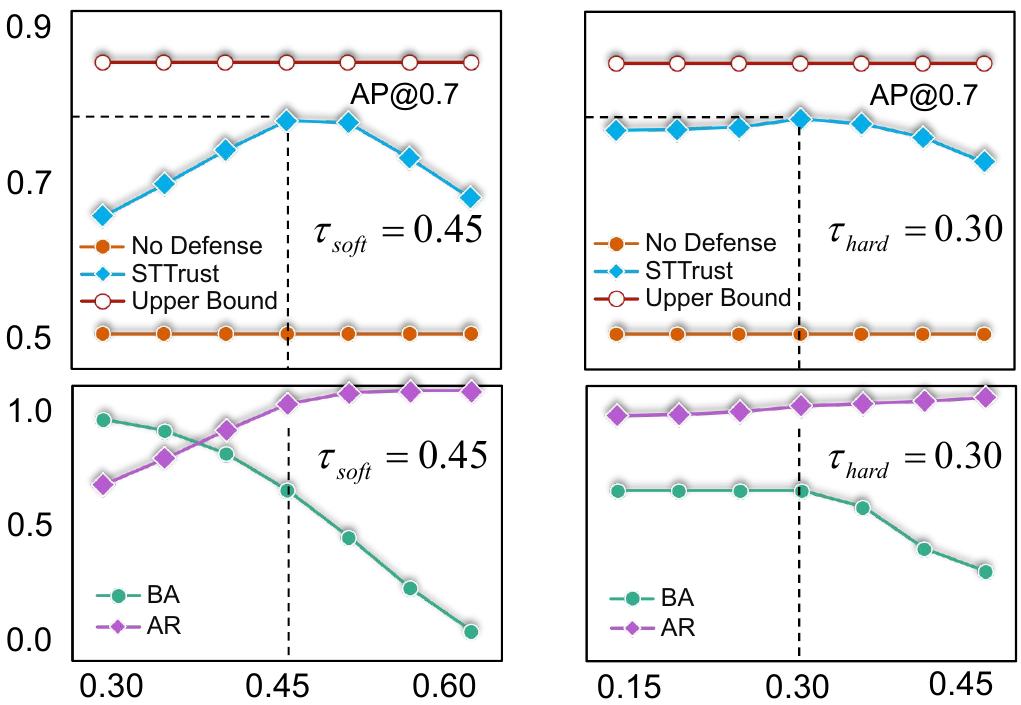}
    \caption{STTrust spatio-temporal consistency threshold analysis on OPV2V~\cite{opv2v} using PointPillars~\cite{pointpillars} with V2VNet~\cite{v2vnet} as the victim model.}
   \label{fig:sttrust_ablation_para}
\end{figure}

\subsection{Hyperparameter Analysis}
\label{sec:Hyperparameter_Analysis}

\textit{PB Confidence Thresholds:}
We investigate the sensitivity of PB to its high- and low-confidence thresholds, as shown in Fig.~\ref{fig:pb_ablation_para} (a, b). By varying one threshold while fixing the other, we find that $\mu_h=0.7$ and $\mu_l=0.3$ provide the best balance between attack effectiveness and defense-bypass rate under FLD~\cite{pb}. This result supports the selected default thresholds for PB on OPV2V~\cite{opv2v}.

\textit{TrPB Surrogate Model Number:}
We investigate the effect of surrogate model pool size on TrPB, as shown in Fig.~\ref{fig:pb_ablation_para} (c). As the number of surrogate models increases, attack effectiveness generally improves, demonstrating the benefit of aggregating knowledge across multiple surrogates for transferable attacks. For each pool size other than five, we evaluate five sampled surrogate pools to account for variations in pool composition. These results support the use of a larger surrogate pool to improve attack transferability.

\textit{STTrust Spatio-Temporal Consistency Thresholds:}
We investigate the sensitivity of STTrust to its spatio-temporal consistency thresholds, as shown in Fig.~\ref{fig:sttrust_ablation_para}. As the soft threshold $\tau_{\mathrm{soft}}$ increases, AP@0.7 initially improves but declines beyond 0.45, while AR increases at the cost of decreasing BA. Varying the hard threshold $\tau_{\mathrm{hard}}$ shows that AP@0.7 peaks around 0.30, with AR remaining high and BA beginning to decrease at larger thresholds. Therefore, $\tau_{\mathrm{soft}}=0.45$ and $\tau_{\mathrm{hard}}=0.30$ provide a favorable balance between perception performance, attack rejection, and benign collaborator acceptance.

\begin{table}[t]
\caption{Ablation of STTrust components against PB, TrPB, and RT-TrPB on OPV2V~\cite{opv2v} using PointPillars~\cite{pointpillars} with V2VNet~\cite{v2vnet} as the victim model. AR and BA denote attack rejection rate and benign collaborator acceptance rate, respectively.}
\centering
\setlength{\tabcolsep}{2pt}
\resizebox{\columnwidth}{!}{
\begin{tabular}{l|ccc|ccc|ccc}
\hline
\multirow{2}{*}{Defense} & \multicolumn{3}{c|}{PB Attack} & \multicolumn{3}{c|}{TrPB Attack} & \multicolumn{3}{c}{RT-TrPB Attack} \\
 & AP Recovery & AR & BA & AP Recovery & AR & BA & AP Recovery & AR & BA \\ \hline
FCV & 7.2/8.2/9.0 & 13.3 & 100.0 & 0.0/0.0/0.1 & 0.1 & 99.8 & 0.3/0.3/0.3 & 0.4 & 98.3 \\
FCV+MTC(B) & 86.2/89.3/89.7 & 90.1 & 70.9 & 77.5/84.9/85.8 & 91.7 & 70.8 & 67.2/72.8/72.7 & 80.9 & 70.9 \\
FCV+MTC & 69.5/70.3/72.6 & 80.3 & 95.9 & 50.8/52.1/54.5 & 50.3 & 95.8 & 22.6/27.4/30.0 & 38.5 & 94.0 \\
STTrust & 95.4/94.6/93.5 & 99.9 & 69.7 & 91.0/90.3/90.1 & 100.0 & 69.7 & 77.0/81.5/81.0 & 95.1 & 69.8 \\ \hline
\end{tabular}
}
\label{tab:sttrust_module_ablation}
\end{table}

\subsection{Qualitative Comparison of Attacks and Defenses}
\label{sec:Visualization_Comparison}
Fig.~\ref{fig:qualitative_main} presents a qualitative comparison of different attacks and defenses on OPV2V~\cite{opv2v} using PointPillars~\cite{pointpillars} with V2VNet~\cite{v2vnet} as the victim model. AdvGPS~\cite{advgps} and James' Attack~\cite{James_attack} produce numerous false positives and false negatives due to their lack of explicit collaboration-consistency preservation, making them more susceptible to consistency-based defenses. In contrast, PB, TrPB, and RT-TrPB maintain partial consistency with ego predictions while selectively manipulating the True Ambiguous Region (TAR) and False Ambiguous Region (FAR), suppressing true detections and reinforcing or fabricating false predictions. The enlarged regions illustrate how these targeted manipulations impair cooperative perception while preserving benign-like consistency. Under PB attacks, STTrust produces results closest to the Upper Bound, substantially reducing false positives and false negatives compared with FLD~\cite{pb} and GLST~\cite{GLST}.

Further implementation details, robustness evaluations of STTrust under challenging conditions, additional TrPB aggregation analysis, random-seed stability analysis, and supplementary quantitative and qualitative visualizations are provided in the supplementary material.

\section{Conclusion}
\label{sec:conclusion}

This work presents a unified framework for stealthy attacks and defense in cooperative perception, built upon the fundamental insight of verification asymmetry. Starting from white-box Pretend Benign (PB), we extend stealthy attacks to zero-query black-box settings through transferable attack generation (TrPB) and further improve their efficiency with real-time generation (RT-TrPB). Complementing these attacks, we develop a spatio-temporal trust-based defense (STTrust) that leverages temporal consistency and historical identity evidence to identify malicious collaborators. Extensive experiments on OPV2V, V2XSet, and V2V4Real demonstrate the effectiveness and stealthiness of the proposed attacks, their transferability across heterogeneous perception models, and the efficiency of real-time attack generation. The results also show that the proposed defense improves perception recovery against both white-box and black-box attacks. This work establishes a systematic foundation for understanding and improving the security of cooperative perception under realistic and heterogeneous threat scenarios.

\bibliographystyle{IEEEtran}
\bibliography{ref}

\newcommand{\compactbio}[3]{%
\par\noindent
\begin{minipage}[t]{0.62in}
\vspace{0pt}\centering
\includegraphics[width=0.56in,height=0.70in]{#1}
\end{minipage}\hspace{0.06in}%
\begin{minipage}[t]{\dimexpr\linewidth-0.72in\relax}
\vspace{0pt}\footnotesize\setlength{\baselineskip}{8.2pt}\textbf{#2} #3
\end{minipage}\par\vspace{5pt}}

\compactbio{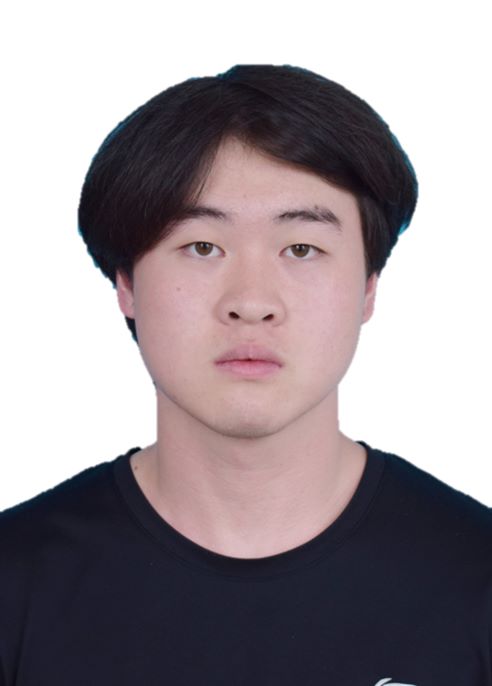}{Hongwei Lin}{is currently pursuing a Ph.D. degree jointly at Xiamen University, Xiamen, China, and Zhongguancun Academy, Beijing, China. His research interests include collaborative perception, adversarial attacks, and embodied intelligence.}

\compactbio{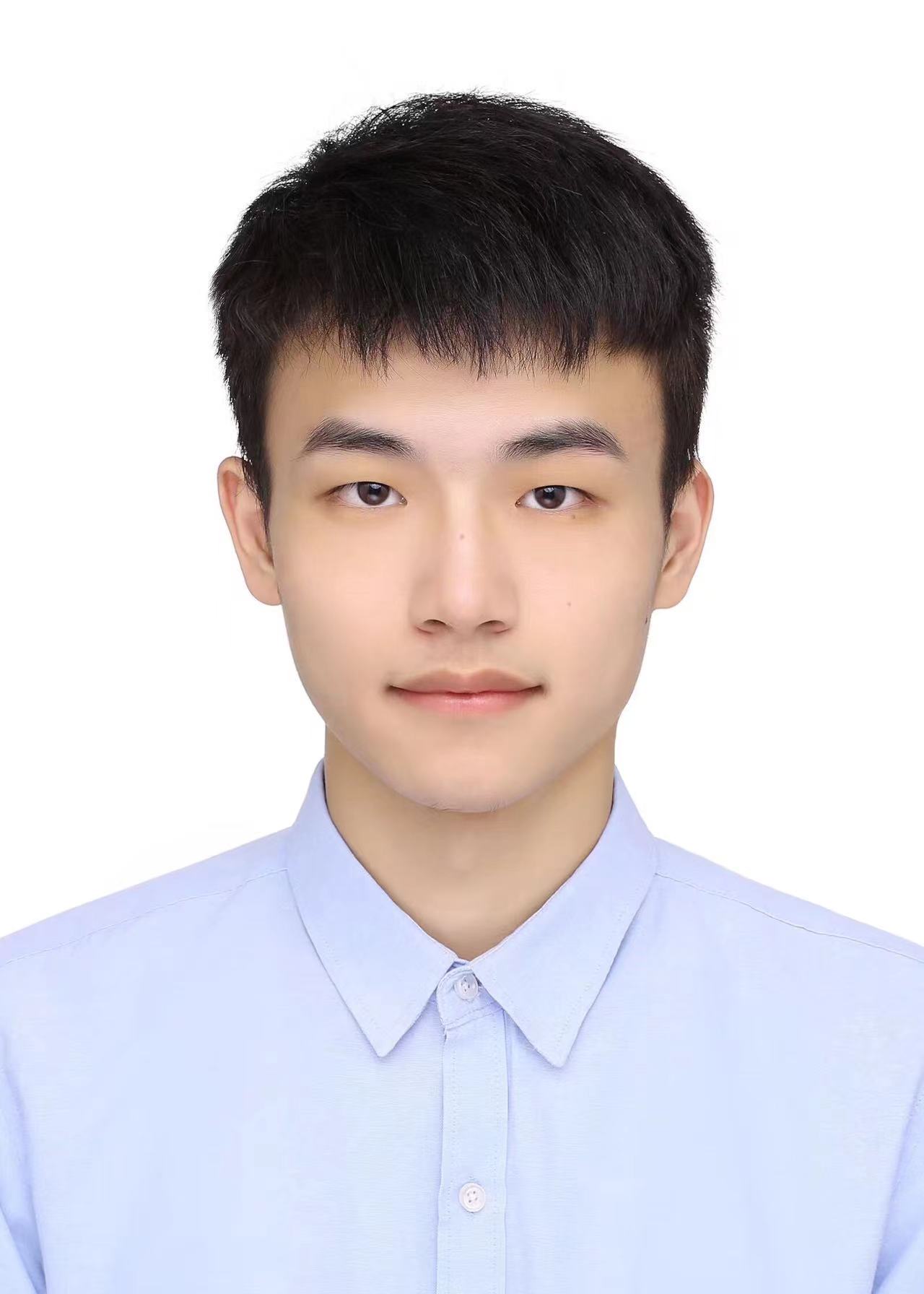}{Xun Huang}{is currently pursuing a Ph.D. degree with the School of Informatics at Xiamen University, Xiamen, China. His research focuses on perception for autonomous driving in adverse weather conditions and embodied active perception. }

\compactbio{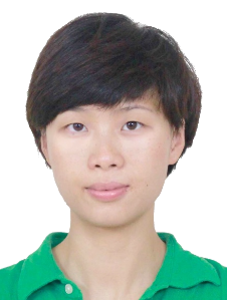}{Chenglu Wen}{received her Ph.D. degree from China Agricultural University, Beijing, China, in 2009. She is currently a professor with the School of Informatics, Xiamen University, Xiamen, China. Her research interests include 3D point cloud processing, 3D vision, and intelligent robots. She is currently a Senior Editor of IEEE TITS and an Associate Editor of IEEE GRSL.}

\compactbio{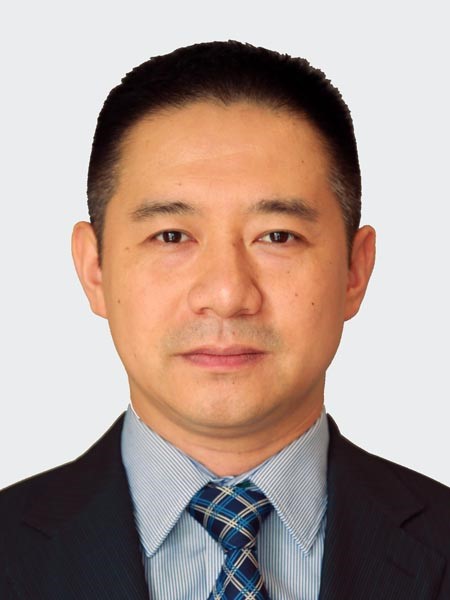}{Cheng Wang}{received his Ph.D. degree from the National University of Defense Technology, Changsha, China, in 2002. He is currently a professor with the School of Informatics, Xiamen University, Xiamen, China. His research interests include image processing, mobile LiDAR data analysis, and multi-sensor fusion. He was the Chair of the ISPRS Working Group I/6 on Multi-sensor Integration and Fusion from 2016 to 2020.}

\end{document}


\title{Zero-Query Transferable Pretend Benign Attacks with Amortized Online Generation \\ for Cooperative Perception}


\maketitle

\appendices

\section{More Experimental and Configuration Details}
\label{sec:Details}

\subsection{Model-specific PB confidence thresholds}
\label{sec:supp_pb_thresholds}

The PB region-selection rule in Pretend Benign uses a high-confidence
threshold $\mu_h$ and a low-confidence threshold $\mu_l$. Because confidence
distributions differ across backbone--fusion configurations, we use
profile-specific values rather than one global threshold. The six profiles
used in the experiments are listed in Table~\ref{tab:pb_thresholds}.

\begin{table}[t]
\caption{Model-specific confidence thresholds for PB region selection.}
\centering
\setlength{\tabcolsep}{20pt}
\resizebox{0.7\columnwidth}{!}{
\begin{tabular}{lcc}
\hline
Configuration & $\mu_h$ & $\mu_l$ \\
\hline
PointPillars + V2VNet   & 0.70 & 0.30 \\
PointPillars + AttFusion & 0.50 & 0.20 \\
PointPillars + F-Cooper  & 0.80 & 0.20 \\
SECOND + V2VNet          & 0.70 & 0.25 \\
SECOND + AttFusion       & 0.70 & 0.20 \\
SECOND + F-Cooper        & 0.90 & 0.20 \\
\hline
\end{tabular}
}
\label{tab:pb_thresholds}
\end{table}

\begin{table}[t]
\caption{STTrust robustness against temporally smoothed white-box PB attacks on OPV2V~\cite{opv2v} using PointPillars~\cite{pointpillars} with V2VNet~\cite{v2vnet} as the victim model.}
\centering
\setlength{\tabcolsep}{4pt}
\resizebox{\columnwidth}{!}{
\begin{tabular}{ccccc}
\hline
\multirow{2}{*}{Temporal weight $\lambda_{\mathrm{temp}}$} & \multicolumn{2}{c}{AP @0.7} & \multirow{2}{*}{Attacker rejection (\%)} & \multirow{2}{*}{Cosine distance} \\
\cmidrule(lr){2-3}
& No Defense     & STTrust    &                                          &                                  \\ \hline
0.00                                       & 0.42           & 0.77       & 99.38\%                                  & 0.06                             \\
0.25                                       & 0.38           & 0.77       & 99.13\%                                  & 0.02                             \\
0.50                                       & 0.38           & 0.76       & 98.79\%                                  & 0.02                             \\
1.00                                       & 0.38           & 0.76       & 98.54\%                                  & 0.02                             \\ \hline
\end{tabular}
}
\label{tab:sttrust_temporal_smoothing}
\end{table}

\subsection{Leave-one-out victim--surrogate protocol}
\label{sec:supp_loo_protocol}

The six configurations are formed by two detection backbones and three fusion
modules. In each transfer experiment, one configuration is held out as the
victim and the remaining five configurations are used as surrogates. The
held-out victim is excluded from surrogate inference, threshold calibration,
shared perturbation optimization, FRN supervision, and RPD construction.

\subsection{Data Splits and Evaluation Isolation}
\label{sec:supp_data_splits}
Perception models are trained on the official training split. TrPB teacher
masks and shared perturbations are generated from this split, which is also
used to train FRN and construct RPD. No frame used for threshold calibration,
FRN fitting, or RPD construction is included in the reported evaluation set.

\section{Robustness of STTrust under Challenging Conditions}
\label{sec:Robustness_of_STTrust}

\begin{table}[t]
\caption{STTrust robustness against intermittent PB attacks on OPV2V~\cite{opv2v}, averaged over six victim models.}
\centering
\setlength{\tabcolsep}{4pt}
\resizebox{\columnwidth}{!}{
\begin{tabular}{ccccc}
\hline
\multirow{2}{*}{Attack pattern} & \multirow{2}{*}{Attacked frames (\%)} & \multicolumn{2}{c}{AP @0.7} & \multirow{2}{*}{Attacker rejection (\%)} \\
\cmidrule(lr){3-4}
&                                       & No Defense     & STTrust    &                                          \\ \hline
attack every frame              & 100.00\%                              & 0.33           & 0.78       & 99.85\%                                  \\
3 on, 2 off, repeat             & 60.05\%                               & 0.68           & 0.78       & 100.00\%                                 \\
1 on, 1 off, repeat             & 50.00\%                               & 0.71           & 0.78       & 100.00\%                                 \\
1 on, 2 off, repeat             & 33.38\%                               & 0.75           & 0.78       & 100.00\%                                 \\
1 on, 4 off, repeat             & 20.11\%                               & 0.79           & 0.78       & 100.00\%                                 \\ \hline
\end{tabular}
}
\label{tab:sttrust_intermittent_pb}
\end{table}

\begin{table}[t]
\caption{STTrust robustness under communication impairments on OPV2V~\cite{opv2v} using PointPillars~\cite{pointpillars} with V2VNet~\cite{v2vnet} as the victim model.}
\centering
\setlength{\tabcolsep}{3pt}
\resizebox{\columnwidth}{!}{
\begin{tabular}{ccccc}
\hline
\multicolumn{1}{c}{\multirow{2}{*}{\begin{tabular}[c]{@{}c@{}}Communication \\ impairment\end{tabular}}} & \multicolumn{2}{c}{AP@0.7} & \multirow{2}{*}{\begin{tabular}[c]{@{}c@{}}Benign \\ Acceptance (\%)\end{tabular}} & \multirow{2}{*}{\begin{tabular}[c]{@{}c@{}}Attacker \\ Rejection (\%)\end{tabular}} \\
\cmidrule(lr){2-3}
\multicolumn{1}{c}{} & Upper Bound & STTrust Defense & & \\ \hline
No impairment & 0.80 & 0.77 & 72.55\% & 100.00\% \\
Position noise $\sigma$ = 0.20 m & 0.59 & 0.59 & 4.40\% & 100.00\% \\
Position noise $\sigma$ = 0.50 m & 0.57 & 0.57 & 0.00\% & 100.00\% \\
Position noise $\sigma$ = 1.00 m & 0.57 & 0.57 & 0.00\% & 100.00\% \\
Yaw noise $\sigma$ = 0.50° & 0.57 & 0.57 & 0.10\% & 100.00\% \\
Yaw noise $\sigma$ = 1.00° & 0.57 & 0.57 & 0.05\% & 100.00\% \\
Yaw noise $\sigma$ = 2.00° & 0.57 & 0.57 & 0.00\% & 100.00\% \\
Packet loss = 10.00\% & 0.80 & 0.75 & 58.30\% & 100.00\% \\
Packet loss = 20.00\% & 0.78 & 0.72 & 45.33\% & 100.00\% \\
Packet loss = 30.00\% & 0.74 & 0.69 & 34.40\% & 100.00\% \\
Fixed delay = 1 frame & 0.75 & 0.73 & 72.08\% & 100.00\% \\
Fixed delay = 2 frames & 0.64 & 0.64 & 71.90\% & 100.00\% \\ \hline
\end{tabular}
}
\label{tab:sttrust_communication_robustness}
\end{table}

\subsection{Robustness to Temporally Smoothed PB Attacks}
We evaluate STTrust against temporally smoothed white-box PB attacks~\cite{pb}, as shown in Table~\ref{tab:sttrust_temporal_smoothing}. We introduce a temporal consistency loss during PB generation to encourage adjacent-frame perturbations to remain similar after ego-pose alignment. As $\lambda_{\mathrm{temp}}$ increases, the cosine distance between aligned adjacent-frame perturbations decreases from 0.06 to 0.02, indicating improved directional consistency. Nevertheless, STTrust maintains an AP@0.7 of 0.76--0.77 and an attack rejection rate above 98.5\% across all settings. These results demonstrate STTrust's robustness to this temporal smoothing strategy, without implying equivalent robustness to all attacks designed to enhance spatio-temporal consistency.

\subsection{Robustness to Intermittent PB Attacks}
We evaluate STTrust against intermittent PB attacks with different attack frequencies on OPV2V~\cite{opv2v}, averaging results over six victim models, as shown in Table~\ref{tab:sttrust_intermittent_pb}. As the proportion of attacked frames decreases from 100\% to 20.11\%, the no-defense AP@0.7 increases from 0.33 to 0.79, reflecting the reduced cumulative impact of intermittent attacks. In contrast, STTrust consistently maintains an AP@0.7 of 0.78 and an attack rejection rate of at least 99.85\% across all attack patterns. These results demonstrate that STTrust remains effective even when PB attacks are intermittently activated.

\subsection{Robustness to Communication Impairments}
We evaluate STTrust under communication impairments, including position and yaw noise, packet loss, and transmission delay, as shown in Table~\ref{tab:sttrust_communication_robustness}. STTrust maintains an attack rejection rate of 100\% across all settings and matches the corresponding Upper Bound in AP@0.7 under position and yaw noise and a two-frame delay. Although benign collaborator acceptance drops sharply under pose noise, the matching AP@0.7 indicates that the corrupted collaborative information provides little additional perception benefit, making its rejection reasonable. Under packet loss and a one-frame delay, STTrust remains within 0.02--0.06 AP@0.7 of the Upper Bound. These results demonstrate robust attack rejection while preserving perception performance close to the achievable bound under communication impairments.

\section{Additional Analysis}
\label{sec:Additional_Analysis}

\begin{table*}[ht]
\caption{White-box attack performance of PB and existing methods across heterogeneous cooperative perception victim models. AP is evaluated at IoU thresholds of 0.3, 0.5, and 0.7 under no defense, ROBOSAC~\cite{robosac}, and FLD~\cite{pb} defenses.}
\centering
\setlength{\tabcolsep}{4pt}
\resizebox{\textwidth}{!}{
\begin{tabular}{ccccccccc}
\hline
\multirow{3}{*}{Victim Model} & \multirow{3}{*}{Attack} & \multirow{2}{*}{No Defense} & \multicolumn{3}{c}{\multirow{2}{*}{ROBOSAC~\cite{robosac} Defense}} & \multicolumn{3}{c}{\multirow{2}{*}{FLD~\cite{pb} Defense}} \\
 & & & \multicolumn{3}{c}{} & \multicolumn{3}{c}{} \\
 \cmidrule(lr){3-3}\cmidrule(lr){4-6}\cmidrule(lr){7-9}
 & & AP@(0.3 / 0.5 / 0.7) & AP@(0.3 / 0.5 / 0.7) $\downarrow$ & $\Delta$AP@(0.3 / 0.5 / 0.7) $\uparrow$ & DBR $\uparrow$ & AP@(0.3 / 0.5 / 0.7) $\downarrow$ & $\Delta$AP@(0.3 / 0.5 / 0.7) $\uparrow$ & DBR $\uparrow$ \\
\hline
\multirow{6}{*}{\begin{tabular}[c]{@{}c@{}}PointPillars~\cite{pointpillars} + \\ AttFusion~\cite{opv2v}\end{tabular}} & No Attack & 0.90 / 0.90 / 0.79 & - & - & - & - & - & - \\
 & PGD~\cite{pgd} & 0.48 / 0.27 / 0.04 & 0.80 / 0.75 / 0.55 & 0.10 / 0.15 / 0.24 & 11.10\% & 0.88 / 0.87 / 0.74 & 0.02 / 0.03 / 0.05 & 0.00\% \\
 & C\&W~\cite{cw-l2} & 0.48 / 0.28 / 0.06 & 0.72 / 0.63 / 0.38 & 0.19 / 0.27 / 0.41 & 30.50\% & 0.88 / 0.87 / 0.74 & 0.02 / 0.03 / 0.05 & 0.00\% \\
 & James' Attack~\cite{James_attack} & 0.62 / 0.37 / 0.09 & 0.79 / 0.74 / 0.48 & 0.11 / 0.16 / 0.31 & 20.80\% & 0.88 / 0.87 / 0.74 & 0.02 / 0.03 / 0.05 & 0.00\% \\
 & AdvGPS~\cite{advgps} & 0.87 / 0.86 / 0.73 & 0.79 / 0.78 / 0.61 & 0.11 / 0.12 / 0.18 & 88.40\% & 0.88 / 0.87 / 0.74 & 0.02 / 0.03 / 0.05 & 0.00\% \\
 & PB (Ours) & 0.65 / 0.62 / 0.52 & \textbf{0.48 / 0.46 / 0.34} & \textbf{0.42 / 0.44 / 0.45} & \textbf{99.70\%} & \textbf{0.61 / 0.60 / 0.51} & \textbf{0.29 / 0.30 / 0.28} & \textbf{100.00\%} \\ \hline
\multirow{6}{*}{\begin{tabular}[c]{@{}c@{}}SECOND~\cite{second} + \\ V2VNet~\cite{v2vnet}\end{tabular}} & No Attack & 0.94 / 0.94 / 0.91 & - & - & - & - & - & - \\
 & PGD~\cite{pgd} & 0.83 / 0.80 / 0.47 & 0.89 / 0.88 / 0.78 & 0.05 / 0.06 / 0.14 & 15.90\% & 0.91 / 0.91 / 0.88 & 0.02 / 0.03 / 0.04 & 0.00\% \\
 & C\&W~\cite{cw-l2} & 0.82 / 0.80 / 0.46 & 0.89 / 0.88 / 0.79 & 0.04 / 0.05 / 0.13 & 16.10\% & 0.91 / 0.91 / 0.88 & 0.02 / 0.03 / 0.04 & 0.00\% \\
 & James' Attack~\cite{James_attack} & 0.90 / 0.89 / 0.52 & 0.90 / 0.89 / 0.84 & 0.04 / 0.04 / 0.07 & 0.80\% & 0.91 / 0.91 / 0.88 & 0.02 / 0.03 / 0.04 & 0.00\% \\
 & AdvGPS~\cite{advgps} & 0.66 / 0.66 / 0.54 & 0.67 / 0.66 / 0.55 & 0.27 / 0.27 / 0.37 & 99.30\% & 0.91 / 0.91 / 0.88 & 0.02 / 0.03 / 0.04 & 0.00\% \\
 & PB (Ours) & 0.33 / 0.33 / 0.28 & \textbf{0.33 / 0.33 / 0.28} & \textbf{0.60 / 0.61 / 0.63} & \textbf{100.00\%} & \textbf{0.34 / 0.33 / 0.28} & \textbf{0.60 / 0.61 / 0.63} & \textbf{99.60\%} \\ \hline
\end{tabular}
}
\label{tab:whitebox_attack_models}
\end{table*}

\subsection{White-Box Attacks Across Heterogeneous Victim Models}
Table~\ref{tab:whitebox_attack_models} evaluates PB across two heterogeneous victim configurations. PB consistently achieves near-perfect defense-bypass rates under both ROBOSAC~\cite{robosac} and FLD~\cite{pb} while substantially degrading perception performance. In contrast, although AdvGPS~\cite{advgps} bypasses ROBOSAC effectively, its DBR drops to 0\% under FLD, as do those of the other baselines. These results demonstrate PB's effectiveness in maintaining stealthy attacks across different perception architectures and fusion strategies.

\begin{table}[t]
\caption{Ablation of TrPB region-specific aggregation strategies on OPV2V~\cite{opv2v} using PointPillars~\cite{pointpillars} with V2VNet~\cite{v2vnet} as the victim model.}
\centering
\setlength{\tabcolsep}{4pt}
\resizebox{\columnwidth}{!}{
\begin{tabular}{cccccc}
\hline
\multirow{2}{*}{Aggregation Strategy} & \multicolumn{2}{c}{ROBOSAC~\cite{robosac} Defense} & \multicolumn{2}{c}{FLD~\cite{pb} Defense} & \multirow{2}{*}{\begin{tabular}[c]{@{}c@{}}Attack Area \\ Ratio (\%)\end{tabular}} \\
\cmidrule(lr){2-3}\cmidrule(lr){4-5}
& $\Delta$AP@0.7      & DBR           & $\Delta$AP@0.7    & DBR         &                                                                                    \\ \hline
TrPB proposed                         & 0.31                & 100.00\%      & 0.31              & 99.79\%     & 1.84                                                                               \\
HCR = majority                        & 0.29                & 100.00\%      & 0.29              & 99.69\%     & 1.96                                                                               \\
HCR = union                           & 0.27                & 100.00\%      & 0.27              & 99.79\%     & 2.08                                                                               \\
TAR = intersection                    & 0.16                & 100.00\%      & 0.16              & 100.00\%    & 1.24                                                                               \\
TAR = union                           & 0.47                & 99.64\%       & 0.24              & 50.44\%     & 2.93                                                                               \\
FAR = intersection                    & 0.29                & 100.00\%      & 0.28              & 98.35\%     & 1.65                                                                               \\
FAR = majority                        & 0.29                & 100.00\%      & 0.29              & 98.30\%     & 1.66                                                                               \\
TIR = intersection                    & 0.25                & 100.00\%      & 0.25              & 99.79\%     & 1.13                                                                               \\
TIR = majority                        & 0.27                & 100.00\%      & 0.27              & 99.85\%     & 1.27                                                                               \\ \hline
\end{tabular}
}
\label{tab:trpb_region_aggregation_ablation}
\end{table}

\subsection{Additional TrPB Aggregation Analysis}

We further analyze the contribution of each region-specific aggregation strategy in TrPB by modifying one region at a time, as shown in Table~\ref{tab:trpb_region_aggregation_ablation}. The proposed configuration achieves the best overall balance, with $\Delta$AP@0.7 of 0.31 under both defenses and an FLD~\cite{pb} DBR of 99.79\%. Specifically, replacing intersection with majority voting or union for HCR reduces attack effectiveness, while intersection for TAR overly restricts attack coverage ($\Delta$AP@0.7=0.16). Conversely, using union for TAR excessively expands the attack region, reducing the FLD DBR to 50.44\%. Replacing union with intersection or majority voting for FAR and intersection or majority voting for TIR also weakens attack effectiveness or defense-bypass performance. These results validate the proposed region-specific aggregation strategy, which preserves reliable consensus for HCR, identifies stable uncertain targets for TAR, and maintains sufficient attack coverage for FAR and TIR.

\section{Random-Seed Stability Analysis}

\begin{table}[t]
\caption{Three-seed stability of white-box PB on OPV2V~\cite{opv2v} for the PointPillars + V2VNet victim model under ROBOSAC~\cite{robosac} and FLD~\cite{pb} defenses.}
\centering
\setlength{\tabcolsep}{4pt}
\resizebox{\columnwidth}{!}{
\begin{tabular}{cccccccl}
\hline
\multirow{2}{*}{seed} & \multirow{2}{*}{No Defense} & \multicolumn{3}{c}{ROBOSAC~\cite{robosac} Defense} & \multicolumn{3}{c}{FLD~\cite{pb} Defense}    \\
\cmidrule(lr){3-5}\cmidrule(lr){6-8}
&    AP@0.7  & AP@0.7  & $\Delta$AP@0.7  & DBR      & AP@0.7 & $\Delta$AP@0.7 & DBR      \\ \hline
2026                  & 0.419                       & 0.419   & 0.431           & 100.00\% & 0.419  & 0.431          & 100.00\% \\
2027                  & 0.418                       & 0.418   & 0.432           & 100.00\% & 0.418  & 0.432          & 100.00\% \\
2028                  & 0.422                       & 0.422   & 0.428           & 100.00\% & 0.422  & 0.428          & 100.00\% \\ \hline
\end{tabular}
}
\label{tab:pb_seed_stability}
\end{table}

\subsection{White-Box PB Seed Stability}
We evaluate the stability of white-box PB across three random seeds on OPV2V for the PointPillars + V2VNet victim model, as shown in Table~\ref{tab:pb_seed_stability}. The no Defense AP@0.7 varies only from 0.418 to 0.422 across seeds, while $\Delta$AP@0.7 remains within 0.428--0.432 under both ROBOSAC and FLD. Both defenses achieve a DBR of 100\% for all three seeds. These results demonstrate that PB maintains consistent attack effectiveness and defense-bypass performance across different random seeds for the evaluated
victim configuration.

\begin{table}[t]
\caption{Three-seed stability of TrPB across six victim models on OPV2V~\cite{opv2v} under FLD~\cite{pb} defense.}
\centering
\setlength{\tabcolsep}{4pt}
\resizebox{\columnwidth}{!}{
\begin{tabular}{cccc}
\hline
\multirow{2}{*}{Victim Model}                                       & \multicolumn{2}{c}{FLD Defense}                             & \multirow{2}{*}{\begin{tabular}[c]{@{}c@{}}Generation \\ (s/frame)\end{tabular}} \\
\cmidrule(lr){2-3}
& $\Delta$ AP @(0.3/0.5/0.7)              & DBR               &                                                                                  \\ \hline
\begin{tabular}[c]{@{}c@{}}PointPillars +\\  V2VNet\end{tabular}    & 0.25 ± 0.00 / 0.26 ± 0.00 / 0.31 ± 0.00 & 99.86\% ± 0.08\%  &  5.69 ± 0.01                                                                      \\ 
\begin{tabular}[c]{@{}c@{}}PointPillars + \\ AttFusion\end{tabular} & 0.18 ± 0.00 / 0.23 ± 0.00 / 0.31 ± 0.00 & 100.00\% ± 0.00\% &  4.83 ± 0.02                                                                      \\ 
\begin{tabular}[c]{@{}c@{}}PointPillars + \\ F-Cooper\end{tabular}  & 0.37 ± 0.00 / 0.38 ± 0.00 / 0.43 ± 0.00 & 99.57\% ± 0.06\%  &  5.67 ± 0.01                                                                     \\ 
\begin{tabular}[c]{@{}c@{}}SECOND + \\ V2VNet\end{tabular}          & 0.20 ± 0.00 / 0.23 ± 0.00 / 0.48 ± 0.00 & 100.00\% ± 0.00\% &  5.65 ± 0.01                                                                     \\ 
\begin{tabular}[c]{@{}c@{}}SECOND + \\ AttFusion\end{tabular}       & 0.19 ± 0.00 / 0.25 ± 0.00 / 0.35 ± 0.00 & 100.00\% ± 0.00\% &  4.81 ± 0.01                                                                     \\ 
\begin{tabular}[c]{@{}c@{}}SECOND + \\ F-Cooper\end{tabular}        & 0.38 ± 0.00 / 0.39 ± 0.00 / 0.38 ± 0.00 & 100.00\% ± 0.00\% &  5.62 ± 0.04                                                                       \\ \hline
\end{tabular}
}
\label{tab:trpb_seed_stability}
\end{table}

\subsection{TrPB Seed Stability in Offline Mode}
We evaluate the three-seed stability of TrPB across six victim models on OPV2V, as shown in Table~\ref{tab:trpb_seed_stability}. The standard deviations of $\Delta$AP at IoU thresholds of 0.3, 0.5, and 0.7 are at most 0.01 across all victim models, while the mean DBR remains between 99.57\% and 100.00\%. Generation time also exhibits little variation across seeds. These results demonstrate that TrPB maintains stable attack effectiveness, transferability, and generation cost across different random seeds and victim architectures.

\begin{table}[t]
\caption{Three-seed stability of RT-TrPB across six victim models on OPV2V~\cite{opv2v} under FLD~\cite{pb} defense.}
\centering
\setlength{\tabcolsep}{4pt}
\resizebox{\columnwidth}{!}{
\begin{tabular}{cccc}
\hline
\multirow{2}{*}{Victim Model}                                       & \multicolumn{2}{c}{FLD Defense}                             & \multirow{2}{*}{\begin{tabular}[c]{@{}c@{}}Generation \\ (ms/frame)\end{tabular}} \\
\cmidrule(lr){2-3}
& $\Delta$ AP @(0.3/0.5/0.7)              & DBR               &                                                                                  \\ \hline
\begin{tabular}[c]{@{}c@{}}PointPillars +\\  V2VNet\end{tabular}    & 0.30 ± 0.01 / 0.30 ± 0.00 / 0.29 ± 0.00 & 97.60\% ± 0.04\%  & 54.23 ± 3.21                                                                      \\
\begin{tabular}[c]{@{}c@{}}PointPillars + \\ AttFusion\end{tabular} & 0.17 ± 0.00 / 0.18 ± 0.00 / 0.22 ± 0.00 & 100.00\% ± 0.00\% & 53.17 ± 2.94                                                                      \\
\begin{tabular}[c]{@{}c@{}}PointPillars + \\ F-Cooper\end{tabular}  & 0.29 ± 0.00 / 0.41 ± 0.01 / 0.47 ± 0.02 & 99.97\% ± 0.03\%  & 40.35 ± 3.03                                                                      \\
\begin{tabular}[c]{@{}c@{}}SECOND + \\ V2VNet\end{tabular}          & 0.10 ± 0.00 / 0.13 ± 0.00 / 0.26 ± 0.00 & 100.00\% ± 0.00\% & 47.39 ± 1.87                                                                      \\
\begin{tabular}[c]{@{}c@{}}SECOND + \\ AttFusion\end{tabular}       & 0.07 ± 0.00 / 0.15 ± 0.00 / 0.41 ± 0.00 & 100.00\% ± 0.00\% & 45.64 ± 2.32                                                                     \\
\begin{tabular}[c]{@{}c@{}}SECOND + \\ F-Cooper\end{tabular}        & 0.38 ± 0.00 / 0.40 ± 0.00 / 0.49 ± 0.01 & 99.99\% ± 0.01\%  & 43.08 ± 2.74                                                                     \\ \hline
\end{tabular}
}
\label{tab:rttrpb_seed_stability}
\end{table}

\subsection{RT-TrPB Seed Stability}
We further evaluate the three-seed stability of RT-TrPB across six victim models on OPV2V, as shown in Table~\ref{tab:rttrpb_seed_stability}. The standard deviations of $\Delta$AP at all three IoU thresholds remain at or below 0.02, while the mean DBR ranges from 97.60\% to 100.00\%. These results indicate that RT-TrPB maintains consistent attack effectiveness and high defense-bypass rates across random seeds and heterogeneous victim architectures.

\subsection{Additional Qualitative Comparison}

\begin{figure*}[t]
  \centering
  \includegraphics[width=\linewidth]{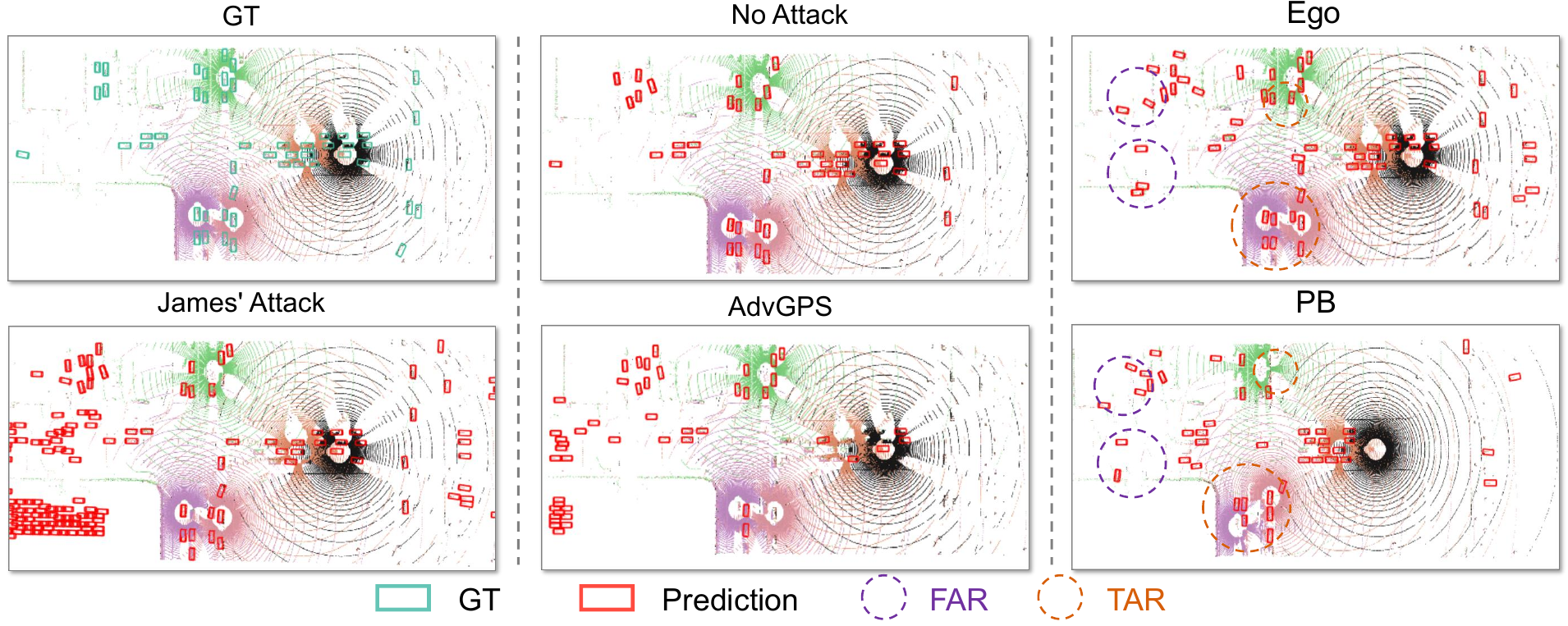}
    \caption{Qualitative comparison of attacks on OPV2V~\cite{opv2v} using PointPillars~\cite{pointpillars} with AttFusion~\cite{opv2v} as the victim model, highlighting the FAR and TAR regions under PB.}
  \label{fig:qualitative_attfusion}
\end{figure*}

\begin{figure*}[t]
  \centering
  \includegraphics[width=\linewidth]{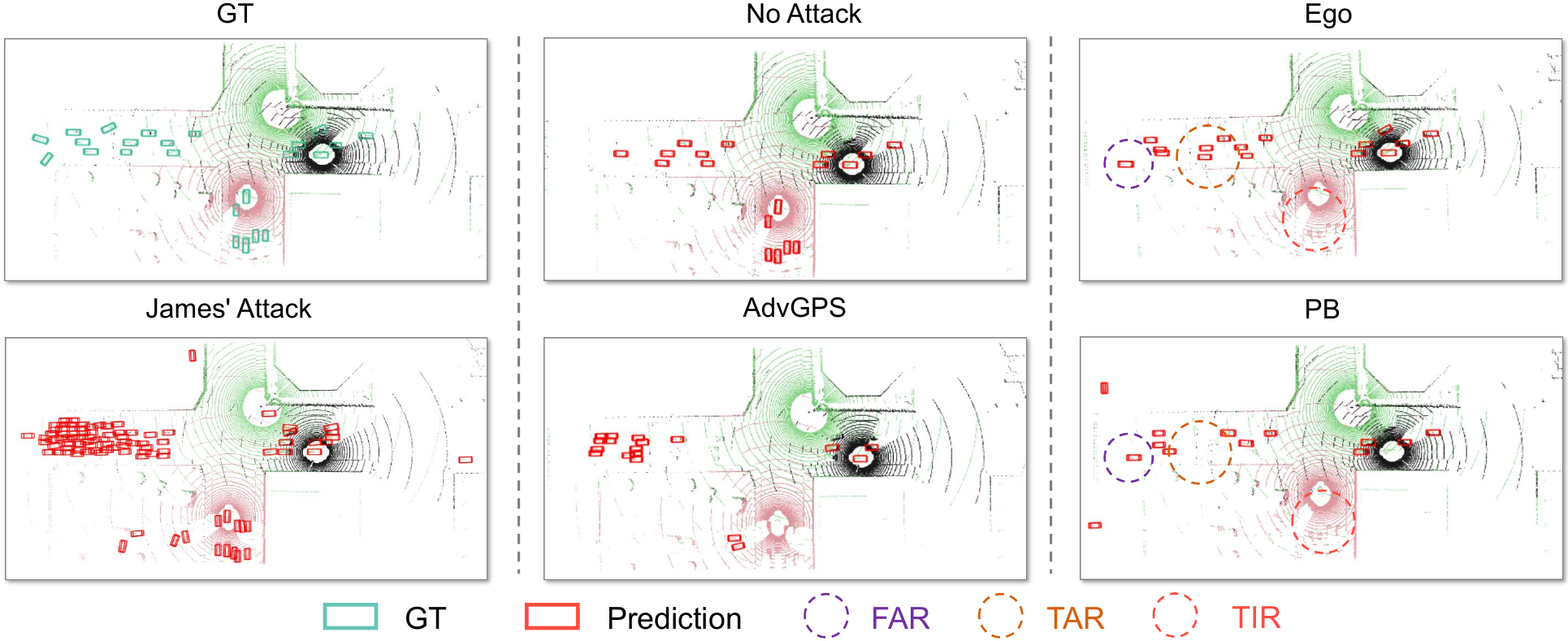}
    \caption{Qualitative comparison of attacks on V2XSet~\cite{v2xset} using PointPillars~\cite{pointpillars} with V2VNet~\cite{v2vnet} as the victim model, highlighting the FAR, TAR, and TIR regions under PB.}
  \label{fig:qualitative_v2xset}
\end{figure*}

\subsection{Cross-Architecture Qualitative Attack Comparison}
\label{sec:Cross-Architecture COMPARISON}
Fig.~\ref{fig:qualitative_attfusion} further visualizes attack outcomes on OPV2V using PointPillars with AttFusion as the victim model. AdvGPS and James' Attack introduce widespread false positives and false negatives, whereas PB preserves partial consistency with ego predictions while selectively manipulating the FAR and TAR regions. These results provide complementary qualitative evidence of PB's region-aware attack behavior under a different fusion architecture.

\subsection{Cross-Dataset Qualitative Attack Comparison}
\label{sec:Cross-Dataset COMPARISON}
Fig.~\ref{fig:qualitative_v2xset} presents complementary attack visualizations on V2XSet~\cite{v2xset} using PointPillars with V2VNet as the victim model. Consistent with the OPV2V results, AdvGPS and James' Attack produce conspicuous false positives and false negatives, while PB preserves partial ego consistency and selectively manipulates the highlighted FAR, TAR, and TIR regions. These visualizations further illustrate PB's region-aware attack behavior on a different dataset.

\bibliographystyle{IEEEtran}
\bibliography{ref}